%% file: Formatting-Instructions-LaTeX-2026.tex
\documentclass[letterpaper]{article} %
\usepackage{aaai2026}  %
\usepackage{times}  %
\usepackage{helvet}  %
\usepackage{courier}  %
\usepackage[hyphens]{url}  %
\usepackage{graphicx} %
\usepackage{natbib}  %
\usepackage{caption} %
\usepackage{algorithm}

\usepackage{algorithmicx}
\usepackage{algpseudocode}
\usepackage{amssymb}
\usepackage{amsmath}
\usepackage{booktabs}
\usepackage{tabularx}

\usepackage{newfloat}
\usepackage{listings}
\DeclareCaptionStyle{ruled}{labelfont=normalfont,labelsep=colon,strut=off} %
\floatstyle{ruled}
\newfloat{listing}{tb}{lst}{}
\floatname{listing}{Listing}
\title{Spherical Geometry Diffusion: Generating High-quality 3D Face Geometry via Sphere-anchored Representations}
\author {
    Junyi Zhang\textsuperscript{\rm 1},
    Yiming Wang\textsuperscript{\rm 1},
    Yunhong Lu\textsuperscript{\rm 1},
    Qichao Wang\textsuperscript{\rm 1},
    Wenzhe Qian\textsuperscript{\rm 1},\\
    Xiaoyin Xu\textsuperscript{\rm 1}, 
    David Gu\textsuperscript{\rm 2},
    Min Zhang\textsuperscript{\rm 1,3,4}\thanks{Corresponding author: Min Zhang (min\_zhang@zju.edu.cn)}
}
\affiliations {
    \textsuperscript{\rm 1}Zhejiang University\\
    \textsuperscript{\rm 2}Stony Brook University\\
    \textsuperscript{\rm 3}Shanghai Institute for Advanced Study-Zhejiang University\\
    \textsuperscript{\rm 4}Shanghai Institute for Mathematics and Interdisciplinary Sciences\\
}

\usepackage{bibentry}

\begin{document}

\maketitle

\begin{abstract}
A fundamental challenge in text-to-3D face generation is achieving high-quality geometry. %
The core difficulty lies in the arbitrary and intricate distribution of vertices in 3D space, making it challenging for existing models to establish clean connectivity and resulting in suboptimal geometry.
To address this, our core insight is to simplify the underlying geometric structure by constraining the distribution onto a simple and regular manifold, a topological sphere.
Building on this, we first propose the Spherical Geometry Representation, a novel face representation that anchors geometric signals to uniform spherical coordinates. This guarantees a regular point distribution, from which the mesh connectivity can be robustly reconstructed. Critically, this canonical sphere can be seamlessly unwrapped into a 2D map, creating a perfect synergy with powerful 2D generative models. We then introduce Spherical Geometry Diffusion, a conditional diffusion framework built upon this 2D map. It enables diverse and controllable generation by jointly modeling geometry and texture, where the geometry explicitly conditions the texture synthesis process.
Our method's effectiveness is demonstrated through its success in a wide range of tasks: text-to-3D generation, face reconstruction, and text-based 3D editing. Extensive experiments show that our approach substantially outperforms existing methods in geometric quality, textual fidelity, and inference efficiency.
\end{abstract}

\section{Introduction}
High-quality 3D face generation is essential for many applications in the computer graphics and movie industry, including virtual reality, computer games, and movie production. Although recent advances have enabled impressive text-driven control, a fundamental challenge persists: achieving high-quality and topologically sound mesh geometry. Many existing methods still produce meshes with significant artifacts and require excessive inference time.%

The ultimate goal of 3D face synthesis is to generate a high-quality triangle mesh, the industry-standard representation renowned for capturing intricate detail. However, the explicit and unstructured nature of the mesh topology makes it exceptionally challenging for generative models to learn directly. This has led to two main approaches, each with significant limitations.
The first approach involves simplifying the problem using the 3D Morphable Model (3DMM)~\cite{wu2023high, aneja2023clipface, kirschstein2024diffusionavatars, zhang2023dreamface, chen2021learning}. By operating in a low-dimensional parametric space, 3DMMs offer easy control but at a steep price: the generated geometry is fundamentally limited by the expressive capacity of the 3DMM prior, precluding high geometric accuracy.
To break free from these constraints, a second and more recent approach uses flexible implicit representations to represent facial geometry~\cite{liu2024headartist, zheng2022imface, zheng2024imface++, yenamandra2021i3dmm, wang2023rodin}. These methods offer a more accurate geometric approximation by learning the signed distance from dense points to the surface. However, to produce a usable mesh, these methods must employ a post-processing algorithm such as Marching Cubes~\cite{lorensen1998marching}. This final step reintroduces the problem of defining \textit{clean vertex connectivity}, often resulting in noisy artifact meshes.
\begin{figure}[t]
    \centering
    \includegraphics[width=0.46\textwidth]{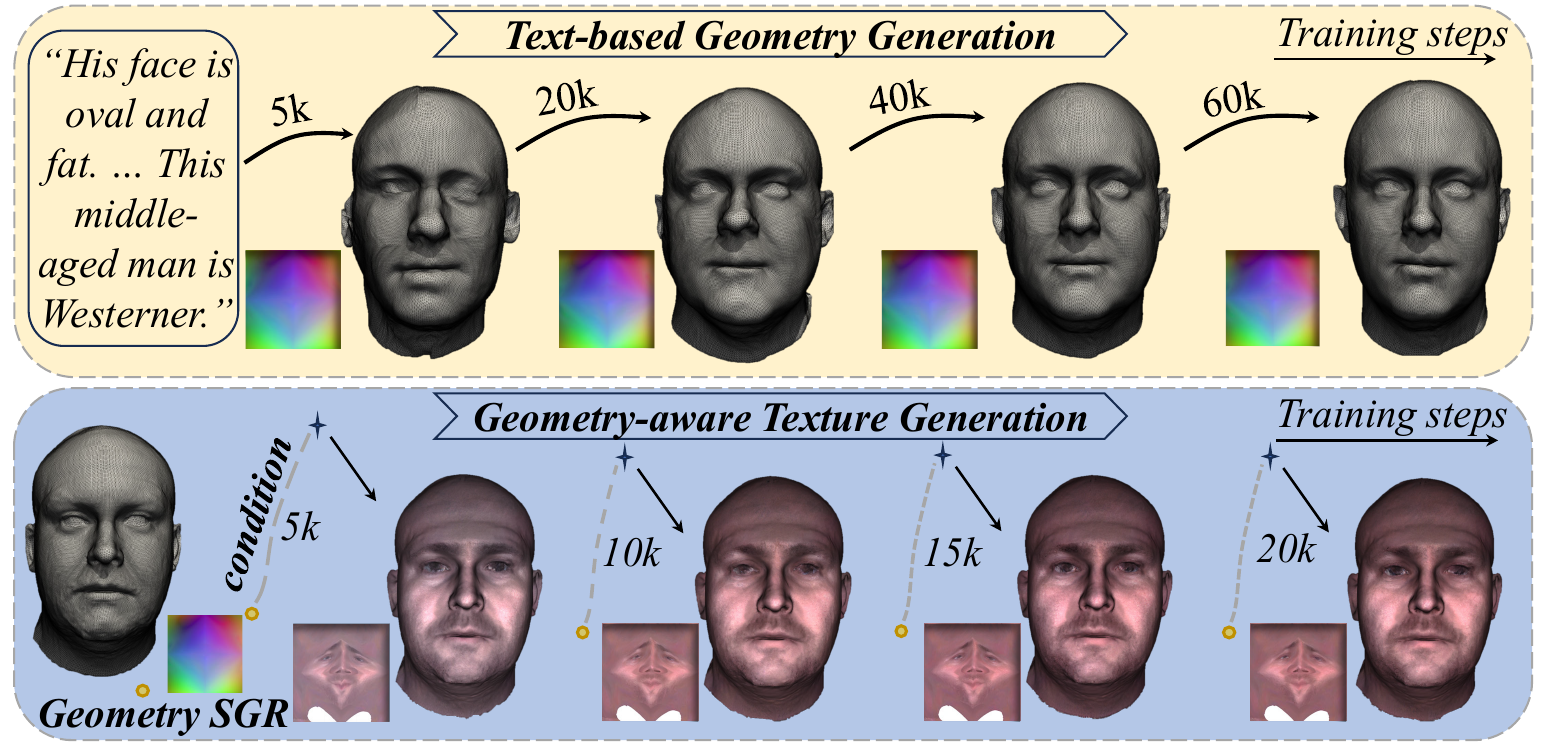}%
    \caption{We present \textbf{Spherical Geometry Diffusion} as a novel framework for 3D faces generation using \textbf{Spherical Geometry Representation}. For \textit{Text-based Geometry Generation}, we achieve flexible control through text conditions. For \textit{Geometry-aware Texture Generation}, we create texture conditioning on the geometry through SGR’s alignment.}
    \label{fig:show_pipeline}
\end{figure}

The \textit{connectivity} problem lies in the arbitrary and unstructured distribution of vertices in the 3D space, where unpredictable spatial relationships make clean topology reconstruction fundamentally challenging.
\textbf{Our key insight is to profoundly simplify this problem by constraining the geometry to a regular manifold}. Instead of learning points in an arbitrary 3D space, we map the facial geometry to a canonical sphere. By anchoring geometric signals to uniformly distributed spherical coordinates, connectivity between points becomes trivial and robust to establish.
This strategy has a two-fold advantage. First, it efficiently solves the connectivity problem. The spherical anchoring inherently defines the mesh topology, eliminating complex extraction and ensuring a high-quality mesh. Second, and just as importantly, this canonical sphere can be seamlessly unwrapped into a 2D map. This creates a structured grid-like domain for the geometry, unlocking the potential to leverage powerful 2D generative models.

Building on this insight, we propose \textbf{Spherical Geometry Representation (SGR)}. This representation parameterizes the 3D face onto a sphere and then unwraps it into a unified 2D map, where each pixel corresponds to a point signal on the sphere. Crucially, this unified map is versatile, capable of encoding not only the 3D vertex positions (geometry), but also the corresponding texture and fine-grained displacement in a unified format. The nature of this map provides an implicit definition of vertex connectivity, allowing a high-quality mesh to be reconstructed by efficient Delaunay spherical triangulation~\cite{fortune2017voronoi}. %

\begin{figure}[t]
    \centering
    \includegraphics[width=0.44\textwidth]{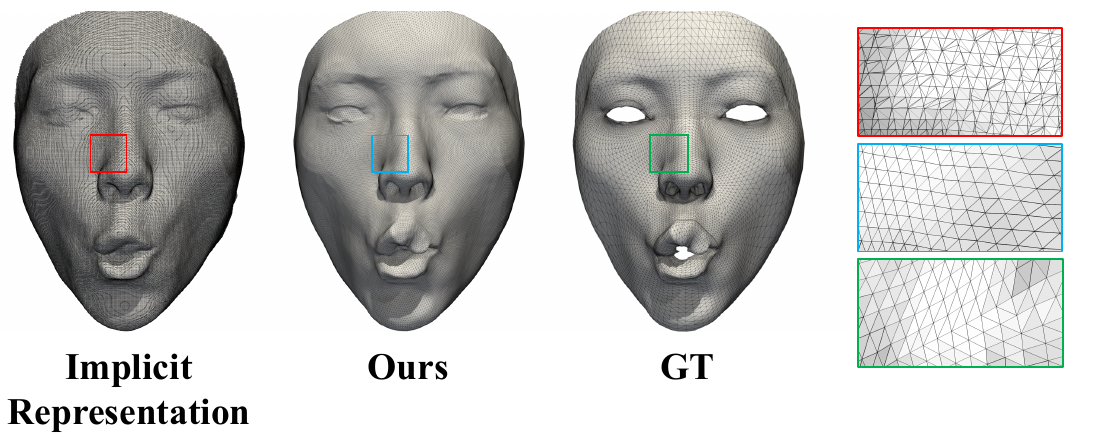}
     \caption{Qualitative comparison of reconstructed mesh quality, where the mesh is reconstructed from the implicit function obtained by ImFace++. A magnified view of the marked region is shown on the right.}
    \label{fig:mesh_rec_compare}
\end{figure}

With geometry and texture elegantly aligned in corresponding 2D structures, we introduce \textbf{Spherical Geometry Diffusion}, a framework designed to efficiently generate high-quality facial geometry, as demonstrated in Fig.~\ref{fig:show_pipeline} and Fig.~\ref{fig:mesh_rec_compare}. Our core innovation is a latent diffusion model that synthesizes a complete geometry in a single forward pass. This native approach stands in stark contrast to conventional text-to-3D methods that require iterative optimization and grants our framework a significant advantage in efficiency.
The unified nature of SGR enables seamless extension to diverse applications beyond pure geometry generation. The framework naturally supports geometry-aware texture synthesis by directly conditioning on the geometry through SGR's shared spherical coordinates, bypassing the complex depth rendering required by traditional pipelines. Furthermore, the architecture enables new face reconstruction and intuitive text-based editing, making it a comprehensive solution for 3D face manipulation.

Beyond the core framework, our work introduces two technical innovations that ensure geometric integrity and training stability. First, to preserve the topology, we develop a novel \textbf{Center-symmetric Padding} scheme. This technique respects the inherent spatial continuity of SGR, effectively eliminating the geometric cracks and boundary artifacts caused by standard 2D convolutions. Second, we propose \textbf{Geometric Regularization} to stabilize the training process. Our regularization leverages SGR's efficient reconstruction properties to provide direct geometric feedback during training, ensuring that the learned latent representations correspond to valid and high-quality facial geometries. This geometric grounding significantly improves both the training efficiency and the final output quality.

We conducted comprehensive experiments on large-scale datasets, evaluating our performance across text-to-geometry generation, face reconstruction, and geometry-aware texture synthesis. The results demonstrate that our method achieves superior performance across all evaluations, producing high-quality geometry that closely matches text prompts and delivering high visual quality.

\begin{figure*}[t]
    \centering
    \includegraphics[width=\textwidth]{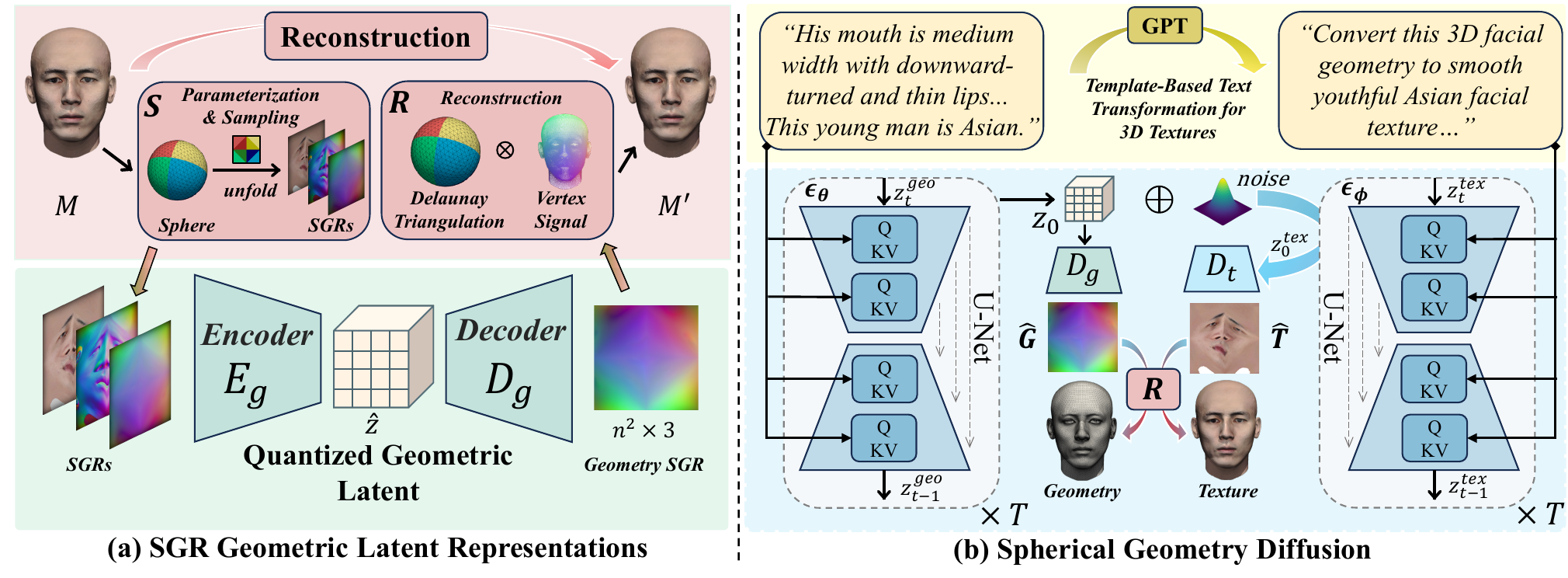} %
    \caption{\textbf{Overview}. Our pipeline comprises two stages. \textbf{(a)} In the first stage, we construct and compress the \textit{\textbf{Spherical Geometry Representation}} into \textit{\textbf{Geometric Latent Representations}}. \emph{Geometric Regularization} and \textit{Center-symmetric Padding} are introduced to enhance geometric quality and accelerate convergence. \textbf{(b)} Using these compact latent space, we train a conditional diffusion model through two sequential phases: (1) \emph{Text-based Geometry Generation}, which generates 3D faces from text prompts, and (2) \emph{Geometry-aware Texture Generation}, which creates textures conditioned on both text and the geometry SGR. The final latent codes are then decoded to reconstruct 3D meshes of high-quality.}
    \label{fig:overview}
\end{figure*}
In summary, our main contributions are as follows.
\begin{itemize}
\item We propose \emph{Spherical Geometry Representation}, which enables faithful bidirectional mapping between 3D and 2D spaces, allowing us to leverage mature diffusion models for the 3D face while ensuring high-quality geometry.
\item We introduce \emph{Spherical Geometry Diffusion}, an efficient framework for controllable 3D face generation, taking advantage of SGR to simultaneously achieve high-quality geometry and texture alignment.
\item We develop two technical innovations, \emph{Center-symmetric Geometric Padding} and \emph{Geometric Regularization}, which ensure geometric quality and stable training.
\end{itemize}

\section{Related Work}
\noindent \textbf{3DMM-based Face Generation.} 
3DMM-based Face Generation focuses on learning the mapping between conditions and 3DMM parameters~\cite{wu2025ice,zhuang2025idol}. The Describe3D method, proposed by Wu~\cite{wu2023high}, learns a mapping function that translates text information in natural language into 3DMM parameters. Many approaches use trained 3DMM parameters as the initial state for a face model and refine geometric details through Score Distillation Sampling (SDS)~\cite{poole2022dreamfusion}. Methods such as FaceG2E~\cite{wu2024text}, DreamFace~\cite{zhang2023dreamface}, HumanNorm~\cite{huang2024humannorm}, and AvatarCraft~\cite{jiang2023avatarcraft} treat the parametric model as the starting geometric configuration and employ SDS techniques to improve the geometry. This improvement is achieved by incorporating 2D depth information and performing a coarse-to-fine optimization process to refine geometric details. \textbf{Spherical Geometry Diffusion}, on the other hand, learns directly from the original 3D data, allowing it to achieve high-quality geometry without the limitation imposed by 3DMM.

\noindent \textbf{Implicit Function for Face Reconstruction.} 
Implicit representations offer an alternative to traditional irregular mesh representations. These representations model 3D shapes by learning continuous deep implicit functions, which can capture shapes at any resolution by querying the occupancy value of points. This approach holds significant promise for high-precision modeling. In recent years, implicit neural representations have been incorporated into 3D face modeling. For example, i3DMM~\cite{yenamandra2021i3dmm} was the first implicit representation model specifically designed for human faces, although its reconstruction accuracy remains suboptimal. Subsequent work, including ImFace~\cite{zheng2022imface}, ImFace++~\cite{zheng2024imface++}, and other recent approaches~\cite{giebenhain2023learning, hong2022headnerf}, has further explored implicit representations for 3D face modeling.%
Despite these advancements, implicit neural representations still depend on post-processing to generate explicit meshes, which results in suboptimal geometric quality. \textbf{Spherical Geometry Diffusion} addresses these challenges by producing high-quality explicit meshes through efficient transformations, ensuring superior geometry.%

\section{Method}
Given a text description, our task is to generate high-fidelity 3D facial meshes.
Our innovations are a novel representation for 3D face and modified Latent Diffusion Models (LDMs) tailored for efficient generation.
The overview of the proposed method is illustrated in Fig.~\ref{fig:overview}. 
The following sections detail our approach.
We begin by introducing \emph{SGR} for representing 3D facial geometry and texture. %
We then describe our Vector Quantized Variational Autoencoder (VQVAE) tailored for geometry SGR. %
Finally, we present a conditional diffusion strategy for generating high-quality 3D faces.%

\subsection{Preliminaries}
\noindent \textbf{Latent Diffusion Models.} Diffusion models are powerful generative models that learn to synthesize data by reversing a fixed forward process that incrementally adds Gaussian noise $\epsilon$ to an input $x_0$. A noisy sample $x_t$ at any time step $t$ can be directly formulated as $x_t = \sqrt{\bar{\alpha}_t}x_0 + \sqrt{1 - \bar{\alpha}_t}\epsilon$, where $\bar{\alpha}_t$ follows a predefined noise schedule. The reverse process is driven by a network $\epsilon_\theta(x_t, t)$, which is trained to predict the noise $\epsilon$ from $x_t$.

LDMs~\cite{rombach2022high} perform this entire diffusion process in a compressed latent space to mitigate the prohibitive computational cost. An autoencoder first maps the data $x$ to a latent code $z_0 = \mathcal{E}(x)$. For conditional synthesis, text embedding $C$ is injected into the denoising U-Net $\epsilon_\theta$ through cross attention. This leads to the objective:
\begin{equation}
    \label{eq:ldm}
    \mathcal{L}_{\text{LDM}} = \mathbb{E}_{\mathcal{E}(x), C, \epsilon \sim \mathcal{N}(0,1), t} \left[ \left\| \epsilon - \epsilon_\theta (z_t, t, C) \right\|_2^2 \right].
\end{equation}

\subsection{Spherical Geometry Representation}
\label{sec:SphereMesh representation}
To constrain the complex distribution of sampling points in the 3D space, we leverage spherical parameterization and uniform sampling to obtain uniformly distributed sampling points in the sphere.
The sampling point values are derived via barycentric interpolation of different signals.

\smallskip
\noindent \textbf{Parameterization and Sampling.} 
Given a 3D face mesh \( \mathcal{M} \), we first map \( \mathcal{M} \) to a spherical mesh \( \mathcal{S} \) using spherical parameterization \( \psi: \mathcal{M} \rightarrow \mathcal{S} \)~\cite{praun2003spherical}. %
Formally, let \( \mathcal{T} = \{\Delta_k\}_{k=1}^M \) be the set of triangular faces in \( \mathcal{M} \), where each \( \Delta_k \) is defined by its vertices \( \mathbf{v}_{k1}, \mathbf{v}_{k2}, \mathbf{v}_{k3} \). The spherical parameterization maps each triangle \( \Delta_k \) to a spherical triangle \( \Delta_k' \) on \( \mathcal{S} \) with vertices \( \mathbf{s}_{k1}, \mathbf{s}_{k2}, \mathbf{s}_{k3} \):
\begin{equation}
    \Delta_k' = \psi(\Delta_k) = \{\psi(\mathbf{v}_{k1}), \psi(\mathbf{v}_{k2}), \psi(\mathbf{v}_{k3})\}, 
\end{equation}
\begin{equation}
    \mathbf{s}_{ki} = \psi(\mathbf{v}_{ki}), \quad i = 1,2,3. \label{eq:vertex_mapping}
\end{equation}
SGR is structured as a 2D image \( \mathbf{G} = \{ \mathbf{G}_{ij} \}_{i=1,j=1}^{H, W} \), where \(W, H\) is the resolution \(R\) of the SGR. We perform uniform sampling in the rectangular domain \( \mathcal{U} \), producing a discrete position \( \mathbf{u}_{ij} = \left( \frac{i}{W}, \frac{j}{H} \right) \) for \( i \in [1, W], j \in [1, H] \). Each sample point \( \mathbf{u}_{ij} \) is then assigned to \( \mathcal{S}\) via an area-preserving bijective mapping \( \phi: \mathcal{U} \rightarrow \mathcal{S} \)~\cite{clarberg2008fast}, resulting in a corresponding position \( \mathbf{s}^{ij} \in \mathbb{R}^3 \) on the sphere:
\begin{equation}
\mathbf{s}^{ij} = \phi(\mathbf{u}_{ij}), \quad \forall i \in [1, W], \ j \in [1, H]. 
\label{eq:area_preserve}
\end{equation}
\( \mathbf{G}_{ij} \) at each position \( \mathbf{u}_{ij} \) represents a weighted surface geometry calculated by barycentric interpolation. Specifically, let \( \mathbf{s}^{ij} \) lie within a spherical triangle \( \Delta_k' \) on \( \mathcal{S} \) with vertices \( \mathbf{s}_{k1}, \mathbf{s}_{k2}, \mathbf{s}_{k3} \), where the signals are \( \mathbf{p}_{k1}, \mathbf{p}_{k2}, \mathbf{p}_{k3} \), respectively. The barycentric coordinates \( (\lambda_0, \lambda_1, \lambda_2) \) of \( \mathbf{s}^{ij} \) with respect to these vertices are:
\begin{equation}
    \lambda_c = \frac{\text{Area}(\mathbf{s}^{ij}, \mathbf{s}_{k(c+1)}, \mathbf{s}_{k(c+2)})}{\text{Area}(\mathbf{s}_{k1}, \mathbf{s}_{k2}, \mathbf{s}_{k3})}, \quad c = 0, 1, 2 .
\end{equation}
where the indices are taken modulo $3$ and \( \text{Area}(\cdot) \) denotes the spherical area formed by the points. With these barycentric coordinates, the interpolated value \( \mathbf{G}_{ij} \) is given by:
\begin{equation}
\mathbf{G}_{ij} = \lambda_0 \mathbf{p}_{k1} + \lambda_1 \mathbf{p}_{k2} + \lambda_2 \mathbf{p}_{k3}.
\label{eq:p}
\end{equation}
It should be noted that signals \( \mathbf{p}_{k1}, \mathbf{p}_{k2}, \mathbf{p}_{k3} \) are optional, allowing SGR to represent not only geometry but also diverse attributes such as texture, colors, and displacement maps through adaptable sampling. Fig.~\ref{fig:exp_example} demonstrates SGR's 2D representation of geometry, textures, and displacement map with the corresponding 3D meshes. 

Furthermore, SGR supports arbitrary resolution. We can improve efficiency or preserve topology by reducing resolution or mapping only the original vertices. Reconstruction results and error at various resolutions are presented in Fig.~\ref{fig:exp_example}. Limiting the mapping to the original vertices preserves the input topology (as shown in Fig.~\ref{fig:exp_example}). In this work, we employ a consistent resolution to align meshes with different vertex counts and ensure a uniform point distribution, which is crucial for high-quality reconstruction.
For details of spherical mapping, please refer to Appendix A.

\smallskip
\noindent \textbf{Mesh Reconstruction.} 
\label{sec:Mesh Reconstruction.}
When reconstructing SGR into an explicit mesh, we first use \( \phi: \mathcal{U} \rightarrow \mathcal{S} \) from Eq. \ref{eq:area_preserve} to map \( \mathbf{G} = \{ \mathbf{G}_{ij} \}_{i=1,j=1}^{H, W} \) to \( \{ \mathbf{S}^{ij} \}_{i=1,j=1}^{H, W}\), and then compute the triangulation on the sphere.
The Delaunay triangulation of spherical points is equivalent to computing the convex hull function $f_{hull}(\cdot)$~\cite{fortune2017voronoi}. Eventually, the triangular face reconstruction \( \mathcal{T''} \) of \( \mathbf{G}  \) is efficiently given by:
\begin{equation}
    \mathcal{T''} = f_{hull}( \{ \mathbf{S}^{ij} \}_{i=1,j=1}^{H, W}).
\end{equation}

\subsection{Geometric Latent Representations}
\label{sec:Geometric Latent Representations}
Being a 2D image structure, our SGR naturally aligns with conditional latent diffusion frameworks. To exploit this property, we first train an encoder \(E_g\) and a decoder \(D_g\) to compress the geometry SGRs into \textbf{Geometric Latent Representations} for effective generation.
Although a standard VQVAE~\cite{esser2021taming} architecture can be applied, we significantly enhance its performance by tailoring the model to the unique properties of our SGR. We introduce two targeted modifications: a novel padding scheme that preserves the spatial continuity of the sphere, and a geometric regularization term that leverages SGR's efficient reconstruction capabilities to accelerate convergence. 
\begin{figure}[t]
    \centering
    \includegraphics[scale=0.4, keepaspectratio]{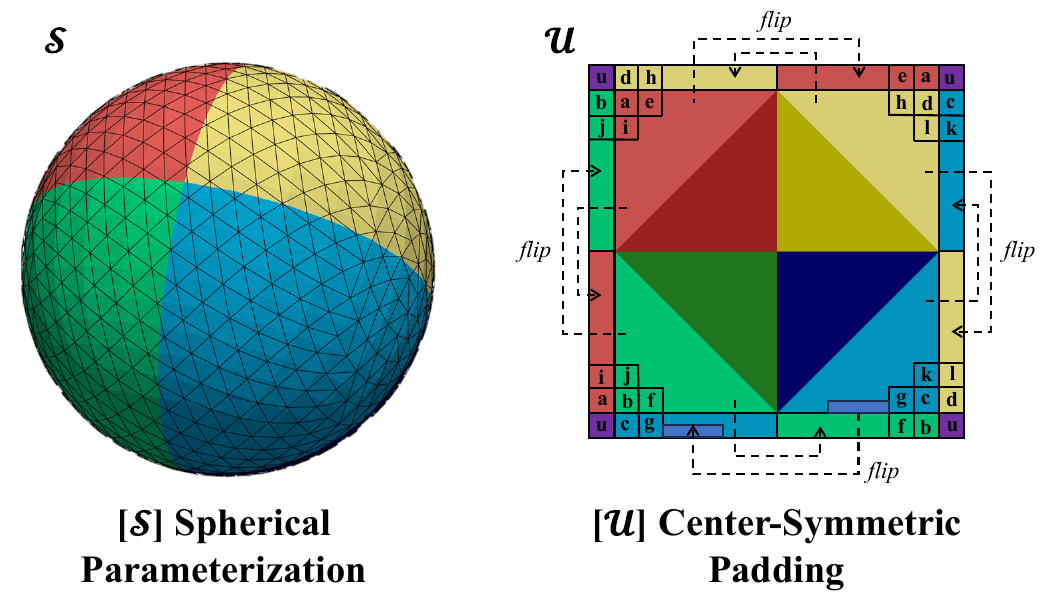} %
    \caption{(\textbf{\( \mathcal{S} \)}) We obtain the spherical domain \( \mathcal{S} \) via Spherical Parameterization \( \mathcal{M} \rightarrow \mathcal{S}\), where (\textbf{\( \mathcal{U} \)}) denotes the unfold SGR grid. Identical colors indicate corresponding regions. (\textbf{\( \mathcal{U} \)}) The proposed \emph{Center-symmetric Padding} mirrors the values symmetrically about the center of each edge. The purple color indicates averaging, with identical values marked by the same letters and colors.}
    \label{fig:spherical}
\end{figure}

\begin{figure}[h]
    \centering
    \includegraphics[scale=0.29, keepaspectratio]{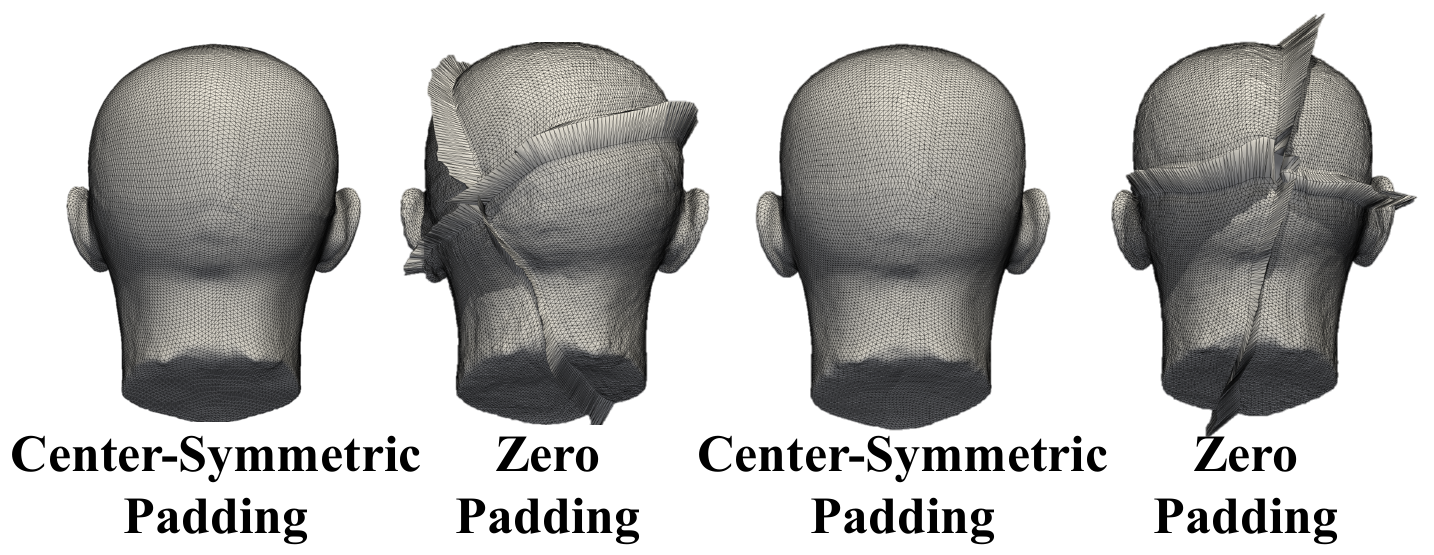} 
    \caption{Impact of \emph{Center-symmetric Padding}.} %
    \label{fig:padding_cut}
\end{figure}

\smallskip
\noindent \textbf{Center-symmetric Padding.} 
\label{sec:Center-symmetric Padding}
When the SGR is unwrapped in a 2D image, it inherently retains spherical continuity. We observed that directly applying traditional zero padding leads to cracks in 3D shapes, as illustrated in Fig.~\ref{fig:padding_cut}. To address this issue, we propose a new padding, \emph{Center-symmetric Padding}, to better adapt to the spherical continuity of the SGR.
Specifically, as illustrated in Fig.~\ref{fig:spherical}, each edge of \( \mathcal{U} \) is centrally symmetric in \( \mathcal{S} \), and the four corners of \( \mathcal{U} \) are averaged in \( \mathcal{S} \). The pseudocode for \textit{Center-symmetric Padding} is shown in Appendix B. %
We use centrally-symmetric pixels as outer padding to convey 3D connectivity, thereby mitigating discontinuities.

\noindent \textbf{Geometric Regularization.} 
\label{sec:Geometric Regularization}
Inspired by VQVAE~\cite{esser2021taming}, we incorporate pixel \( \mathcal{L}_{\text{pix}} \), perceptual \( \mathcal{L}_{\text{per}} \), and adversarial losses \( \mathcal{L}_{\text{adv}} \) in our model. 
The loss function \( \mathcal{L}_{rec} \) is a weighted sum of :
\begin{equation}
    \mathcal{L}_{\text{rec}} = \mathcal{L}_{\text{pix}} + \mathcal{L}_{\text{per}} + \lambda_{\text{adv}} \mathcal{L}_{\text{adv}}
\end{equation}
\( \lambda_{\text{adv}} \) controls the weight of \(\mathcal{L}_{\text{adv}}\). For details of \( \mathcal{L}_{\text{pix}} \), \( \mathcal{L}_{\text{per}} \), and \( \mathcal{L}_{\text{adv}} \), please refer to Appendix B.

However, traditional VQVAE suffers from high training costs and slow convergence. Leveraging the efficient conversion of SGR to 3D meshes, we introduce geometric regularization from a mesh perspective, which accelerates convergence and significantly boosts performance.

Given a reconstructed mesh \( M = (V, F) \), where \( V \) denotes the vertices and \( F \) represents the faces, we define regularization as a weighted sum of the following losses.

\textbf{\textit{Normals consistency loss}}, \( \mathcal{L}_{nor} \), encourages the alignment of the normals of adjacent faces in the mesh. For two neighboring faces \( f_0 \) and \( f_1 \) with respective normals \( n_0 \) and \( n_1 \),  the loss is calculated as:
\begin{equation}
\mathcal{L}_{nor} = \sum_{\langle f_0, f_1 \rangle} \left( 1 - \frac{n_0 \cdot n_1}{\|n_0\| \|n_1\|} \right)
\end{equation}

\textbf{\textit{Laplacian smoothing loss}}, \( \mathcal{L}_{lap} \), promotes smoothness across the mesh by penalizing large variations in vertex positions. It is defined as:
\begin{equation}
\mathcal{L}_{lap} = \sum_{i=1}^{N_v} \left\| L v_i \right\|^2
\end{equation}
where \( N_v \) is the number of vertices in the mesh, and \( L \) is the Laplacian matrix applied to the vertex \( v_i \). %

\textbf{\textit{Edge length regularization loss}}, \(\mathcal{L}_{edge}\), penalizes the formation of distorted vertices. For each edge \( e \) in the mesh, the loss compares its actual edge length \( \| e \| \) to a target length \( e_0 \):\begin{equation}
\mathcal{L}_{edge} = \frac{1}{N_m} \sum_{m=1}^{N_m} \frac{1}{E_m} \sum_{e \in m} \left( \| e \| - e_0 \right)^2
\end{equation}
where \( N_m \) is the number of meshes, \( E_m \) is the number of edges in each mesh, and \( \| e \| \) represents the length of edge \( e \). %

The total loss of our geometric VQVAE is given by
\begin{equation}
\mathcal{L}_{total} = \alpha_{nor} \cdot \mathcal{L}_{nor} + \alpha_{lap} \cdot \mathcal{L}_{lap} + \alpha_{edg} \cdot \mathcal{L}_{edge} + \mathcal{L}_{rec}
\end{equation}
where \( \alpha_{nor}, \alpha_{lap}\), and \( \alpha_{edg} \) are the respective weights for each loss term, allowing controlled emphasis on different aspects of \textit{Geometric Regularization}. %

\subsection{Spherical Geometry Diffusion}
\label{sec:SphereMesh Diffusion}
By representing 3D assets as structured 2D maps, our \textbf{Spherical Geometry Diffusion} bypasses the complexities of 3D convolutions and irregular mesh processing. This reframes 3D generation as a 2D synthesis problem, allowing us to harness the pre-trained weights and optimization strategies of mature image diffusion models to boost performance.

We implement this through a decoupled two-stage process. First, \textit{Text-based Geometry Generation} provides fine-grained control over facial structure from text prompts. Second, \textit{Geometry-aware Texture Generation} leverages SGR's intrinsic alignment, using the geometry map as a perfect pixel-aligned condition for texture synthesis. This strategy guarantees geometric-textural consistency, a common failure in end-to-end models.

\smallskip
\noindent \textbf{Text-based Geometry Generation.} 
To efficiently generate the geometry SGR \( G \) from text embedding $C$, we model the distribution $p(G | C)$ within the latent space of our geometric VAE. %
The geometric UNet $\epsilon_\theta$ is trained to denoise the latent $z_t^{\text{geo}}$, conditioned on the timesteps $t$ and CLIP embeddings $C$ of the text prompt. $C$ is injected through cross-attention to minimize our objective in Eq. \ref{eq:ldm}. At inference time, this trained model reverses the process. Starting from a random Gaussian latent noise, $\epsilon_\theta$ iteratively denoises the latent vector, conditioned on the text embedding $C$. The resulting clean latents are then decoded by the geometric VAE decoder $D_g$ to produce the final geometry SGR $\hat{G}$.

\smallskip
\noindent \textbf{Geometry-aware Texture Generation.} 
For the second stage, we learn to synthesize a corresponding texture SGR \(T\). We frame this as a conditional image-to-image translation task, which is motivated by the intrinsic pixel-level correspondence of SGR. This key decision allows us to bypass depth rendering for alignment and instead leverage the pre-trained image diffusion checkpoint.%

Specifically, we implement this by fine-tuning a texture generator $\epsilon_\phi$, to denoise a latent texture $z_t^{\text{tex}}$. Following InstructPix2Pix~\cite{brooks2023instructpix2pix}, we first distill the comprehensive prompt \(y\) into a texture-specific description \(y_{tex}\), which is then embedded to embedding \(c_{tex}\). This is achieved by template prompts and GPT~\cite{brown2020language}. Crucially, a strong geometric prior is supplied by concatenating the latent representation of the geometry, $E_g(G)$, with the noised texture latent $z_t^{\text{tex}}$ along the channel dimension. We minimize the following latent diffusion objective:
\begin{equation}
\mathcal{L}_{\text{tex}} = \mathbb{E}_{\substack{
    E_t(T), E_g(G), C_{tex}, \\ 
    \epsilon \sim \mathcal{N}(0,1), t
}}
\left[ \| \epsilon - \epsilon_\phi(z_t^{\text{tex}}, t, E_g(G), C_{tex}) \|_2^2 \right].
\label{eq:your_label_here}
\end{equation}

\section{Experimental Results}
\paragraph{Implementation Details.}
We define the geometry SGR resolution ($W, H$) as 256x256, yielding reconstructed meshes with 65,536 vertices. For the corresponding texture SGRs, we employ a higher resolution of 512x512. Geometry SGRs are stored as 16-bit, three-channel PNGs, whereas all other SGRs utilize an 8-bit, three-channel format.
For the text-based geometry generation stage, we train the Latent Diffusion Model for 500,000 steps on four NVIDIA A100 GPUs with a learning rate of $1\text{e-}4$. The geometric regularization loss weights are set to $\alpha_{\text{nor}} = 0.1$, $\alpha_{\text{lap}} = 0.5$, and $\alpha_{\text{edg}} = 0.1$. To ensure training stability, we introduce an adversarial loss ($\lambda_{\text{adv}}=0.1$) after the first 100 epochs. We employ the DPM-Solver++ scheduler~\cite{lu2022dpm} with Classifier-Free Guidance~\cite{ho2022classifier}.
For the geometry-aware texture generation stage, the model is fine-tuned from the official Stable-Diffusion-XL (SDXL) checkpoint~\cite{podell2023sdxl}. Further implementation details are available in Appendix C.%

\paragraph{Datasets.}
We evaluate the performance of our method on various datasets, including Describe3D~\cite{wu2023high}, FaceScape~\cite{yang2020facescape}, and COMA~\cite{ranjan2018generating}. Describe3D contains 1,627 high-quality 3D face meshes paired with fine-grained textual descriptions, covering a wide age range and diverse ethnic backgrounds. FaceScape is a large-scale dataset that contains 16,572 3D scans of 847 individuals, each captured with 20 different expressions. The COMA dataset contains diverse sequential expression sequences from 12 individuals, totaling 17,794 meshes. Since FaceScape and COMA lack fine-grained text descriptions, we constructed structured text based on their attribute annotations, such as age, gender, and expression. As these attributes are not related to appearance, we primarily use these two datasets to evaluate the face geometry.

\subsection{Quantitative Comparison}
\noindent \textbf{Conditional 3D Face Generation.} 
To assess the geometric generation performance of our method, we conduct a quantitative evaluation on the FaceScape and COMA dataset. Following prior work~\cite{taherkhani2023controllable}, we employ two metrics: Specificity (SP), which assesses how closely generated faces fit with training data distribution, and Diversity (DIV), which assesses the variety among generated samples. We compared our method with other mesh generation approaches, including 3DMM~\cite{amberg2008expression}, MAE~\cite{abrevaya2018multilinear}, CoMA~\cite{ranjan2018generating}, FacialGAN~\cite{abrevaya2019decoupled},and D3FSM~\cite{abrevaya2019generative}. Our method demonstrates impressive performance across both metrics,effectively generating diverse and high-fidelity facial geometries, as shown in Table~\ref{tab:gen_comparison}.
Best performance are \textbf{bolded}, second-best are \underline{underlined}.
\begin{table}[t]
    \centering
    \setlength{\tabcolsep}{3pt}
    \small
  \begin{tabular}{
    >{\centering\arraybackslash}m{2 cm} |
    >{\centering\arraybackslash}m{1.5cm} | 
    >{\centering\arraybackslash}m{2.2 cm} |  
    >{\centering\arraybackslash}m{1.5cm} 
  }
        \toprule
        Method & DIV$\uparrow$ & DIV-ID$\uparrow$ & SP (mm)$\downarrow$ \\
        \midrule
        Training data & 1.00 & 1.00 & -    \\
        \midrule
        3DMM & 0.72 & 0.59 & 2.30 \\
        MAE & 0.79 & 0.28 & 2.00 \\
        CoMA & 0.69 & 0.52 & 2.47 \\
        FacialGAN & \textbf{0.96} & 0.58 & 2.01 \\
        D3FSM & 0.77 & \underline{0.81} & \underline{0.84} \\
        Ours & \underline{0.93} & \textbf{0.88} & \textbf{0.82} \\
        \bottomrule
    \end{tabular}
    \caption{Quantitative results of conditional geometry generation with respect to Diversity and Specificity.}
    \label{tab:gen_comparison}
\end{table}
\begin{table}[t]
  \centering
  \small
  \setlength{\tabcolsep}{3pt}
  \begin{tabular}{
    >{\centering\arraybackslash}m{2 cm} |
    >{\centering\arraybackslash}m{1.5cm} | 
    >{\centering\arraybackslash}m{2.2cm} |  
    >{\centering\arraybackslash}m{1.5cm} 
  }
    \toprule
    Method & CD (mm) $\downarrow$ & F-score@1mm $\uparrow$ & AR $\downarrow$\\
    \midrule
    i3DMM & 0.759 & 83.87 & 3.18\\
    FLAME & 0.662 & 88.18 & \underline{1.84}\\
    FaceScape & 0.731 & 79.38 & 1.93\\
    NPHM & 0.637 & 91.67 & 3.90\\ 
    ImFace & 0.553 & 94.67 & 4.10\\
    ImFace++ & \underline{0.511} & \underline{96.11} & 4.34\\
    Ours & \textbf{0.466} & \textbf{98.77} & \textbf{1.35}\\
    \bottomrule
  \end{tabular}
\caption{Quantitative results of 3D face reconstruction.}%
\label{tab:rec_comparison}
\end{table}

\noindent \textbf{Conditional 3D Face Reconstruction.} 
We quantitatively evaluated the reconstruction of 3D face geometry on the FaceScape dataset. Following the official data split~\cite{zheng2022imface}, we train our model and then reconstruct the geometry from the test set using DDIM inversion~\cite{dhariwal2021diffusion}. We benchmark our approach against state-of-the-art (SOTA) methods employing alternate geometric representations, including FLAME~\cite{li2017learning}, FaceScape~\cite{yang2020facescape}, i3DMM~\cite{yenamandra2021i3dmm}, NPHM~\cite{giebenhain2023learning}, ImFace~\cite{zheng2022imface}, and ImFace++~\cite{zheng2024imface++}. We report Chamfer Distance (CD) and F-score as metrics for geometric accuracy, while Aspect Ratio (AR)~\cite{parthasarathy1994comparison} evaluates geometric quality, as shown in Table~\ref{tab:rec_comparison}. %
The results demonstrate that our method achieves superior performance, outperforming all competing baselines in both geometric accuracy and mesh quality.

\noindent \textbf{Conditional Texture Generation.} 
We quantitatively evaluate the visual performance on the Describe3D dataset using a CLIP-based assessment. This involves comparing our method against Describe3D~\cite{wu2023high}, DreamFace~\cite{zhang2023dreamface}, HumanNorm~\cite{huang2024humannorm} and FaceG2E~\cite{wu2024text}, all tested on a shared set of 20 text prompts. In Table~\ref{tab:clip_comparison}, we report the CLIP score~\cite{sanghi2023clip} for text-mesh alignment, inference time (in minutes) for efficiency, and Ranking-1 for the times a method was ranked first according to CLIP. We use the official DreamFace website for its time calculations. The results demonstrate that our method achieves high fidelity and efficiency. As a native 3D generative model, it offers a significant advantage in computational speed over iterative SDS-based methods. %
\begin{figure}[t]
    \centering
    \small
    \includegraphics[width=1.0\columnwidth, keepaspectratio]{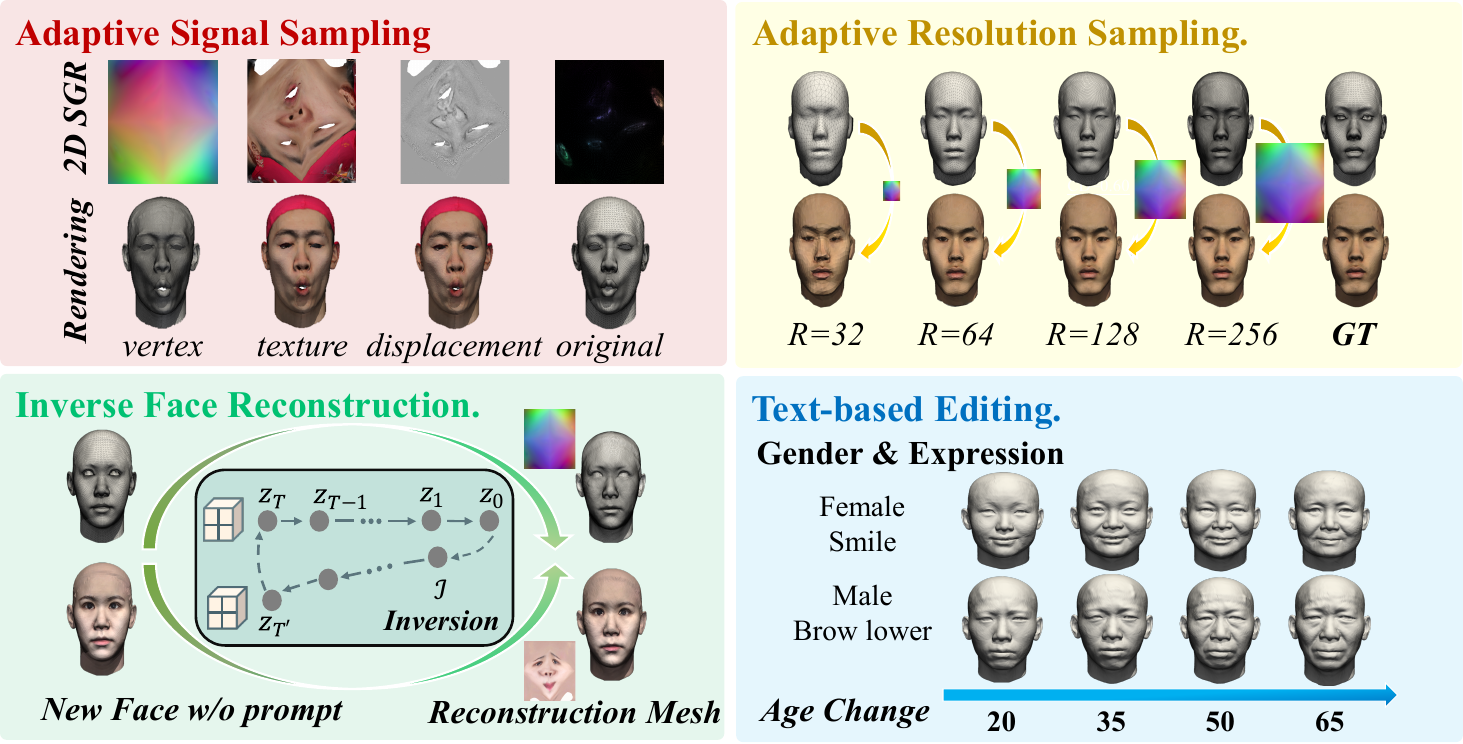}%
    \caption{Our SGR enables adaptive signals and resolution. Leveraging the diffusion framework, we also enable inverse face reconstruction for new faces without prompts, along with text-based editing for gradual identity aging.}%
    \label{fig:exp_example}
\end{figure}
\begin{table}[t]
  \centering
  \small
  \setlength{\tabcolsep}{3pt}
  \begin{tabular}{
    >{\centering\arraybackslash}m{2 cm} |
    >{\centering\arraybackslash}m{1.5cm} | 
    >{\centering\arraybackslash}m{2.2cm} |  
    >{\centering\arraybackslash}m{1.5cm} 
  }
    \toprule
    Method & Score $\uparrow$ & Ranking-1 $\uparrow$ & Time $\downarrow$ \\
    \midrule
    Describe3D & 28.01 & 1 & \underline{0.7 m} \\
    DreamFace & 28.63 & 1 & 1.0 m \\
    HumanNorm & 29.42 & 3 & 85 m \\
    FaceG2E & \underline{29.74} & \textbf{8} & 5.0 m \\
    Ours & \textbf{29.80} & \underline{7} & \textbf{0.2 m} \\
    \bottomrule
  \end{tabular}
\caption{CLIP evaluation results on the synthesized faces. Best performance are bolded, second-best are underlined.}
\label{tab:clip_comparison}
\end{table}

\subsection{Qualitative Comparison}
\begin{figure}[t]
    \centering
    \includegraphics[width=1.0\columnwidth, keepaspectratio]{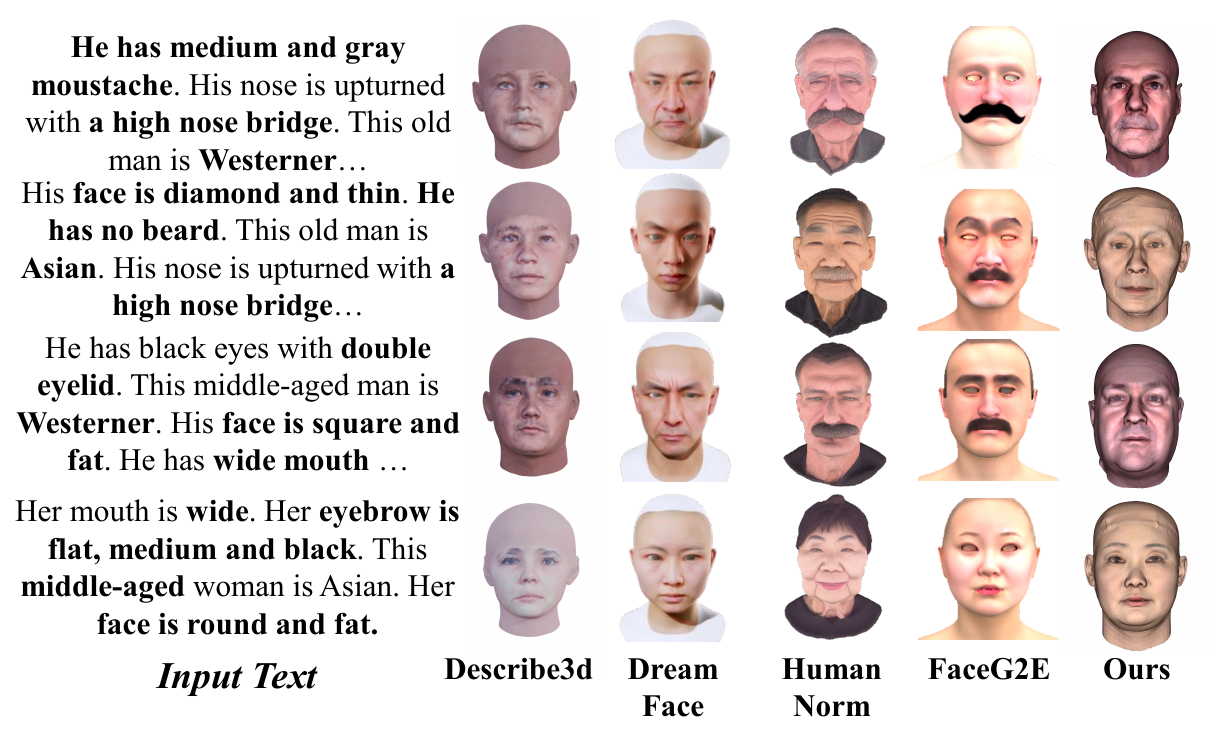} 
    \caption{Qualitative results of Geometry-aware Texture Generation. Bold text indicates key aspects.}%
    \label{fig:mesh_generation}
\end{figure}

\noindent \textbf{Conditional 3D Face Generation.} 
In Fig.~\ref{fig:mesh_generation}, we present qualitative results of our method compared to other methods. The results demonstrate that our method responds better to the fine-grained details of the text prompt.

\begin{figure}[t]
    \centering
    \includegraphics[width=1.0\columnwidth, keepaspectratio]{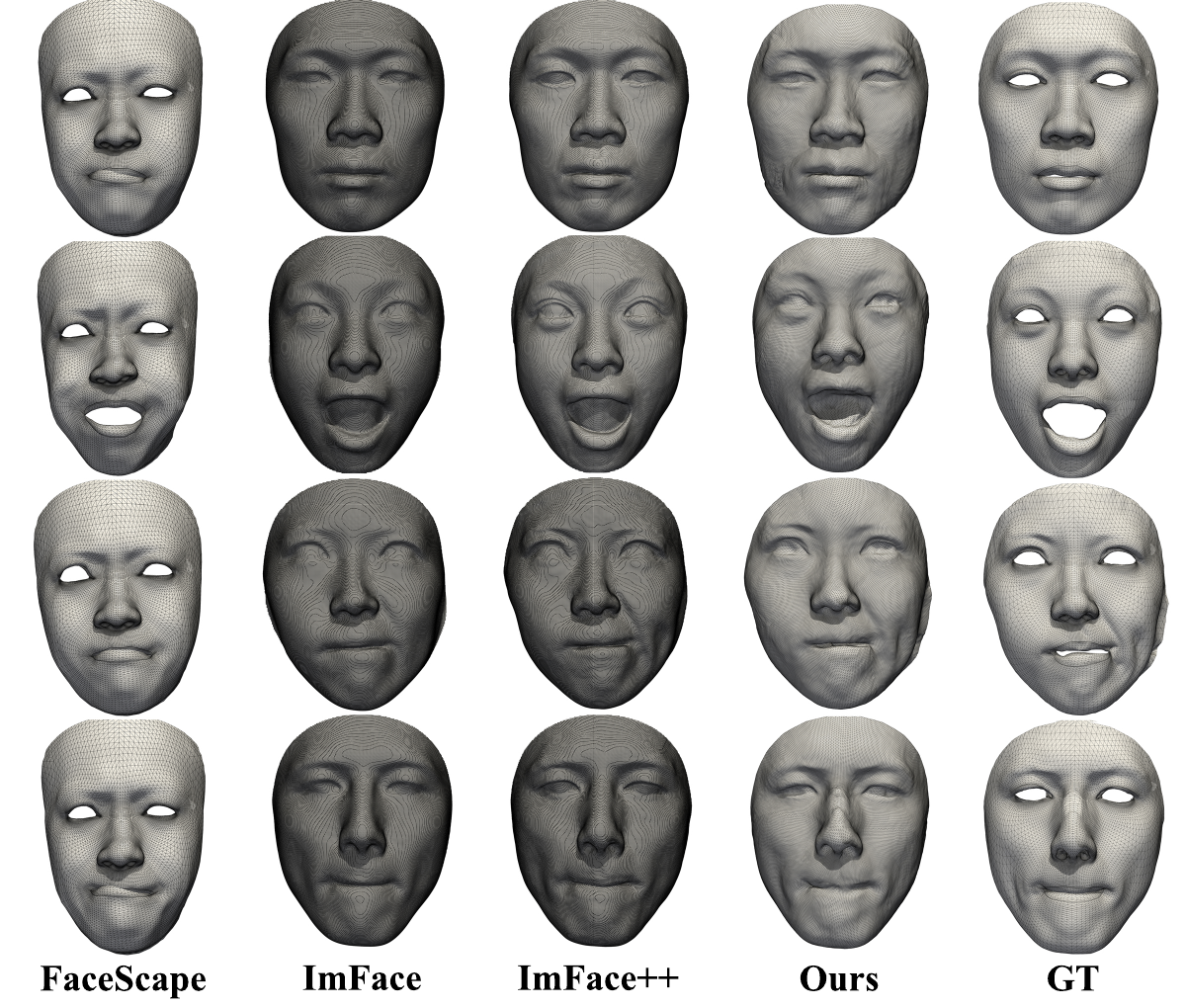} 
    \caption{Qualitative results of 3D Face Reconstruction.}%
    \label{fig:mesh_reconstruction}
\end{figure}

\noindent \textbf{Conditional 3D Face Reconstruction.}
We present qualitative results for Fig.~\ref{fig:mesh_reconstruction}. The results show that our method produces more accurate and detailed reconstructions. Moreover, Fig.~\ref{fig:mesh_rec_compare} shows that our method generates uniform meshes, while the SDF-based methods exhibit noticeable artifacts. %

\noindent \textbf{Further Analysis.}
To demonstrate the versatility of our SGR representation, we conducted experiments on two advanced tasks: direct text-based editing and out-of-domain synthesis using an SDS framework~\cite{poole2022dreamfusion}.
For text-based editing, we show that modifying semantic attributes in the prompt, such as age or expression, enables precise corresponding edits to the facial geometry while preserving personal identity. As demonstrated in Fig.~\ref{fig:exp_example}, increasing progressively the age in text prompt results in more pronounced facial wrinkles. This highlights our method's high degree of control over geometric features and its ability to capture fine details. More qualitative results, including those from our SDS experiments, can be found in Appendix D.

\subsection{Ablation Study}
\noindent \textbf{Center-symmetric Padding.}
We conduct a qualitative ablation study on the effectiveness of \emph{Center-symmetric Padding}. As shown in Fig.~\ref{fig:padding_cut}, \emph{Center-symmetric Padding} helps to eliminate cracks and improve performance.

\noindent \textbf{Geometric Regularization.}
We performed a quantitative experiment comparing model with and without \textit{Geometric Regularization}, as shown in Table~\ref{tab:combined_ablation}. \emph{VQVAE-mesh} (with regularization) and \emph{VQVAE-base} (without regularization) are equally trained for 30,000 steps. The results show that \textit{Geometric Regularization} helps penalize deformed meshes, enabling faster convergence and improving efficiency.

\noindent \textbf{Inversion steps.}
The quality of reconstruction during the inversion process is influenced by the number of steps taken. Generally, more steps lead to better quality, but this also increases computational time. To investigate this trade-off, we conduct ablation experiments. As illustrated in Table~\ref{tab:combined_ablation}, the performance improves with more inversion steps. However, we observed diminishing returns, with additional steps beyond a certain point yielding only marginal gains. Consequently, we chose 600 inversion steps for our experiments to strike a balance between efficiency and quality.

\noindent \textbf{Inference steps and guidance scale.}
In diffusion models, the inference step and the guidance scale are hyperparameters that regulate the generation performance. We conducted ablation studies to examine the impact of these variables. The results of this analysis, along with further ablation experiments, are detailed in Appendix D.
\begin{table}[t]
    \centering
    \small
    \setlength{\tabcolsep}{6pt}
    \begin{tabular}{l|c|c}
    \toprule
    Setting & CD(mm) $\downarrow$ & F-score@1mm $\uparrow$ \\
    \midrule
    \multicolumn{3}{l}{\textit{Inversion Steps}}\\
    \midrule
    \textit{100 steps} & 1.235 & 66.70 \\
    \textit{200 steps} & 0.924 & 81.59 \\
    \textit{300 steps} & 0.693 & 93.00 \\
    \textit{400 steps} & 0.571 & 97.23 \\
    \textit{500 steps} & 0.571 & 97.24 \\
    \textit{\textbf{600 steps}} & \underline{0.465} & \textbf{98.77} \\
    \textit{700 steps} & 0.465 & 98.76 \\
    \textit{800 steps} & 0.465 & 98.77 \\
    \textit{900 steps} & 0.465 & 98.77 \\
    \textit{1000 steps} & \textbf{0.466} & \underline{98.76} \\
    \midrule
    \multicolumn{3}{l}{\textit{Geometric Regularization}}\\
    \midrule
    \emph{VQVAE-base} & 0.810 & 74.83 \\
    \emph{VQVAE-mesh} & \textbf{0.666} & \textbf{85.16} \\
    \bottomrule
    \end{tabular}
    \caption{Ablation study on the number of inversion steps and effectiveness of geometric regularization.}
    \label{tab:combined_ablation}
\end{table}
\section{Conclusion}
\label{sec:Conclusion}
In this paper, we introduced Spherical Geometry Diffusion, a framework that tackles the critical challenge of poor geometric fidelity in text-to-3D face generation. By representing facial structure on a canonical sphere, our method ensures clean mesh topology and seamlessly leverages powerful 2D diffusion models, establishing a new state of the art in geometric quality and textual alignment. We believe this principle of regularizing complex 3D structures onto simple manifolds offers a robust and scalable paradigm for future high-fidelity content creation.

\smallskip
\noindent \textbf{Limitation.}
Although our method marks a significant advance, its performance is strained by data availability. Public text-3D face datasets, constrained by privacy concerns, are much smaller than 2D image datasets. We anticipate performance improvements as this data gap narrows.

\section{Acknowledgments}
This work was supported by the National Major Science and Technology Projects (2022ZD0117000 to M. Zhang), the National Natural Science Foundation of China (62202426 to M. Zhang), and the National Institutes of Health (NIH) (R21EB029733 to X. Gu).

\bibliography{aaai2026}
\input{supp_body}

\end{document}

%% file: supp_body.tex
\begin{figure*}[!ht]
    \centering
    \vspace*{0.5cm}
    {\fontsize{16pt}{23pt}\selectfont\textbf{Spherical Geometry Diffusion: Generating High-quality 3D Face Geometry via Sphere-anchored Representations\\Appendix}}
    \vspace{1cm}
\end{figure*}

\appendix
\setcounter{secnumdepth}{1} 
\section{Background}
\label{Supp:1}
\subsection{Parameterization}
Our spherical parameterization algorithm originates from a stretch-optimized framework, enabling efficient mapping of complex 3D models onto a spherical domain with minimal distortion. The method operates through two key phases: initial spherical parametrization of the input mesh followed by a distortion-minimizing stretch optimization. For comprehensive implementation details, we refer readers to the original work~\cite{praun2003spherical}.

\smallskip
\noindent \textbf{Stretch Efficiency.}
The theoretical framework for stretch analysis originates from~\cite{sander2001texture} in planar domains. Consider a parametrization \( \psi: \mathcal{U} \rightarrow \mathcal{M} \) that maps planar coordinates \( (s, t) \in \mathcal{U} \subset \mathbb{R}^2 \) to 3D points \( (x, y, z) \in \mathcal{M} \subset \mathbb{R}^3 \). The deformation characteristics of this mapping are captured by the \( 3 \times 2 \) Jacobian matrix:
\[
J_\psi = \left( \frac{\partial \psi}{\partial s} \frac{\partial \psi}{\partial t} \right)
\]
Singular values of \( J_\psi \) at any point \( (s,t) \), \( \Gamma \) and \( \gamma \), representing the largest and smallest lengths when transforming unit vectors from \( \mathcal{U} \) to \( \mathcal{M} \). These values derive two scalar stretch norms root-mean-square \( L^2 \) and worst-case \( L^\infty \):
\[
L^2(s, t) = \sqrt{\frac{1}{2}(\Gamma^2 + \gamma^2)} \quad \text{and} \quad L^\infty(s, t) = \Gamma
\]
The \( L^2 \)-stretch metric integrates local distortions across the entire surface:
\[
L^2(\mathcal{M}) = \sqrt{\frac{1}{A_\mathcal{M}} \iint\limits_{(s,t) \in \mathcal{U}} \left(L^2(s, t)\right)^2 dA_\mathcal{M}(s, t)} 
\]
where \( dA_\mathcal{M}(s, t) =  \gamma \Gamma \, ds \, dt \) is the differential surface area element. \textit{stretch efficiency} in triangle is piecewise linear, so the Jacobian \( J_\psi \) is constant over each triangle. Derived from this, the \textit{$L^2$ stretch efficiency} \( \eta = {A_\mathcal{M}}/{A_\mathcal{U}}\cdot{1}/{L^2(\mathcal{M})^2} \) where \( A_\mathcal{M}\) and \( A_\mathcal{U} \) are the areas of the surface and domain respectively. \( \eta \) quantifies parametrization quality, bounded by \( \eta \leq 1 \). 

\smallskip
\noindent \textbf{Spherical Stretch Efficiency.}
Subsequent work~\cite{praun2003spherical} adapts these principles to spherical parametrizations \( \psi: \mathcal{S} \rightarrow \mathcal{M} \), where \( \mathcal{S} \) denotes a spherical domain. The inverse mapping \( \psi^{-1}: \mathcal{M} \rightarrow \mathcal{S} \) provides analytical advantages for distortion analysis. For any triangle \( T \subset \mathcal{M} \), consider local orthonormal coordinates \( (s,t) \) on \( T \) and their spherical counterparts through \( \psi^{-1} \). The Jacobian \( J_{\psi^{-1}} \) of this inverse mapping reveals reciprocal stretch factors \( 1/\gamma \) and \( 1/\Gamma \), where \( \Gamma,\gamma \) are singular values of \( J_{\psi^{-1}} \).
The spherical \( L^2 \)-stretch over the triangle \( T \) thus becomes:
\[
L^2(T) = \sqrt{\frac{1}{A_{\mathcal{M}_T}} \iint\limits_{(s,t) \in T} \left( \frac{1}{\gamma^2} + \frac{1}{\Gamma^2} \right) dA_{\mathcal{M}_T}(s, t)},\quad 
\]
where \( dA_{\mathcal{M}_T}(s,t) = ds \, dt \) is the differential area of the mesh triangle.
To avoid bunched-up and wavy iso-parameter lines when \( \Gamma \gg \gamma \), they add a small regularization term \( \varepsilon (A_\mathcal{M}/4\pi)^{p/2+1} \Gamma^p \) that penalizes inverse stretch. This term effectively mitigates oversampling artifacts without imposing conformal constraints.

\smallskip
\noindent \textbf{Optimization of Stretch Efficiency.}
Minimizing the stretch norm is a nonlinear optimization. To effectively optimize the \textit{stretch efficiency}, the algorithm uses a coarse-to-fine multiresolution strategy. The surface mesh \( \mathcal{M} \) is progressively simplified to a tetrahedron, creating a progressive mesh that favors triangles with good aspect ratios. The base tetrahedron is initially mapped to the sphere, and vertices are incrementally inserted while maintaining an embedding and minimizing the stretch metric.

During the progressive mesh traversal, each vertex split introduces a new vertex whose neighborhood is defined by a ring of existing vertices, forming a spherical polygon. To ensure an embedding and avoid flipped or degenerate triangles, the new vertex must be placed within the kernel of this spherical polygon. The kernel is defined as the intersection of the open hemispheres bounded by the polygon edges, and this kernel is guaranteed to be non-empty if the mapping was an embedding prior to the vertex insertion. 

After inserting a new vertex, local optimization is performed on its neighboring vertices. Each vertex in the neighborhood is optimized individually by minimizing the stretch metric summed over its adjacent triangles. This optimization is achieved through great-circle searches in random directions on the sphere, using bracketed parabolic minimization. To prevent flipping, the vertex is constrained to lie within the kernel of its 1-ring neighborhood. This constraint also avoids degenerate triangles, as they would contribute infinite stretch energy.

Periodically, after the number of vertices has increased by a constant factor, a global optimization sweep is performed. All vertices are placed in a priority queue, ordered by the magnitude of change in their local neighborhoods. The optimization process iterates until the largest change in any neighborhood falls below a predefined threshold.

\subsection{Equal-Area Mapping}
Our \textit{Equal-Area mapping algorithm}, originating from~\cite{clarberg2008fast}, unfolds the unit sphere into a unit square while preserving fractional area. This algorithm offers features such as low distortion, straightforward interpolation, and efficient forward and inverse transforms. We outline the computational framework of this algorithm, with detailed implementation instructions available in the original work~\cite{clarberg2008fast}.

\smallskip
\noindent \textbf{Hemisphere Mapping.}
We first introduce the mapping of the hemisphere, as it forms the basis for the entire sphere mapping.
For mapping the hemisphere, concentric squares are mapped to concentric circles while maintaining fractional area. In the first sector, where \( \phi \in \left[-\frac{\pi}{4}, \frac{\pi}{4}\right] \), a point \( (s, t) \) in the unit square \( \mathcal{U} = [0,1]^2 \) is transformed to a point on the hemisphere \( \mathcal{H} = \{(x, y, z) \mid x^2 + y^2 + z^2 = 1, z \geq 0\} \) using the following steps.

\begin{align}
    (s, t) &\rightarrow
    \begin{cases}
        u = 2s - 1 \\
        v = 2t - 1
    \end{cases} \nonumber \\
    &\rightarrow
    \begin{cases}
        r = u \\
        \phi = \frac{\pi}{4} \frac{v}{u}
    \end{cases} \nonumber \\
    &\rightarrow
    \begin{cases}
        x = \cos \phi \cdot r \sqrt{2 - r^2} \\
        y = \sin \phi \cdot r \sqrt{2 - r^2} \\
        z = 1 - r^2
    \end{cases} \tag{1}\label{eq:1}
\end{align}

Similar transforms apply to the other sectors. In the last step \( z = 1 - r^2 = \cos \theta \), where \( \theta \) is the angle from the \( z \)-axis. Consequently, \( \sin \theta = \sqrt{1 - \cos^2 \theta} = r \sqrt{2 - r^2} \), which explains the expressions for \( x \) and \( y \). Similar transformations are applied to the remaining sectors.

\smallskip
\noindent \textbf{Sphere Mapping.}
To achieve an equal-area mapping of the entire sphere, the unit square is divided into eight triangles. The four innermost triangles map to the northern hemisphere, while the outer four are folded down to cover the southern hemisphere. Each triangle corresponds to a quadrant in one of the two hemispheres. The transformation from the unit square \( \mathcal{U} \) to the sphere \( \mathcal{S} = \{(x, y, z) \mid x^2 + y^2 + z^2 = 1\} \) is derived as follows:

For the innermost triangle in the first quadrant, the lengths \( a \) and \( b \) are given by:
\begin{align}
    a = \frac{u + v}{\sqrt{2}},\quad b = \frac{v - u}{\sqrt{2}}. \nonumber
\end{align}
The transformation to the unit disk is then:
\begin{align}
    r &= \sqrt{2}a = u + v, \nonumber \\
    \phi &= \frac{\pi}{4} \frac{b}{a} = \frac{\pi}{4} \left( \frac{v - u}{r} + 1 \right), \quad \text{with } 0 \leq \phi \leq \frac{\pi}{2}, \tag{2}\label{eq:2}
\end{align}
where \( \phi \) is measured from the positive \( u \)-axis.

For the outermost triangle, which maps to the southern hemisphere, the parameterization is obtained by mirroring the innermost triangle about the diagonal. Here, \( b \) remains \( {(v - u)}/2{\sqrt{2}} \), while \( a \) is given by \( a = {(2 - u - v)}/{\sqrt{2}} \). The transformation to the unit disk is:
\begin{align}
    r &= \sqrt{2}a = 2 - u - v, \nonumber \\
    \phi &= \frac{\pi}{4} \frac{b}{a} = \frac{\pi}{4} \left( \frac{v - u}{r} + 1 \right), \quad \text{with } 0 \leq \phi \leq \frac{\pi}{2}. \tag{3}\label{eq:3}
\end{align}
The mapping from the disk to the sphere follows the same final step as Eq. \ref{eq:1}, except that the \( z \)-component is negated (\( z = -(1 - r^2) \)) for the outer triangle. The remaining quadrants are similarly handled using analogous transformations.

The implementation of this algorithm avoids branching and incorporates other optimizations to accelerate computation. For these details, we refer readers to the wrok~\cite{clarberg2008fast}.

\subsection{3D Face Generation}
\noindent \textbf{Multi-view image.}
Approaches based on 2D images \cite{tran2018nonlinear,tewari2021learning,chen2025rapverse,lin2025towards} typically train models using large datasets of 2D images captured from multiple viewpoints, with these diverse facial images aiding the model in generating a variety of facial shapes. Recently, generative methods, such as diffusion models specifically tailored for face generation, have shown impressive results \cite{chan2021pi}. Inversion-based methods, utilizing generative models for face reconstruction, have also demonstrated their potential by producing adaptable parameters that can be used to reconstruct customized facial data \cite{mokady2023null,lu2025inpo,lu2025smoothed}. While these methods achieve visually compelling results, they face limitations in 3D geometric accuracy, as 2D images lack sufficient geometric detail. Furthermore, these methods often require complex post-processing to produce usable 3D meshes. In contrast, Spherical Geometry Diffusion directly learns from native 3D data, achieving superior 3D generation quality without the need for complex post-processing.

\noindent \textbf{3DMM-based.}
The Describe3D method, proposed by Wu et al. \cite{wu2023high}, learns a mapping function that translates natural language into 3DMM parameters. Many approaches use the trained 3DMM parameters as the initial state for a face model, refining the geometric details through SDS techniques \cite{poole2022dreamfusion}. Methods such as those by Wu et al. \cite{wu2024text}, DreamFace \cite{zhang2023dreamface}, HumanNorm \cite{huang2024humannorm}, and AvatarCraft \cite{jiang2023avatarcraft} treat the parametric model as the starting geometric configuration and employ SDS techniques to enhance the geometry. This enhancement is achieved by incorporating 2D depth information and performing a coarse-to-fine optimization process to refine geometric details. Spherical Geometry Diffusion, on the other hand, directly learns from native 3D data, enabling it to achieve higher geometric accuracy and quality without relying on intermediate parametric representations.

\noindent \textbf{Implicit Neural Representations.}
Implicit representations offer an alternative to traditional irregular mesh representations. These representations model 3D shapes by learning continuous deep implicit functions, which can capture shapes at any resolution by querying the occupancy value of points. This approach holds significant promise for high-precision modeling. In recent years, implicit neural representations have been incorporated into 3D face modeling. For example, i3DMM \cite{yenamandra2021i3dmm} was the first implicit representation model specifically designed for human faces, although its reconstruction accuracy remains suboptimal. Subsequent works, including ImFace \cite{zheng2022imface}, ImFace++ \cite{zheng2024imface++}, and other recent approaches \cite{giebenhain2023learning, hong2022headnerf}, have further explored implicit representations for 3D face modeling, resulting in improvements in reconstruction accuracy. Despite these advancements, implicit neural representations still depend on post-processing to generate display meshes and lack explicit geometric structures, which limits their direct applicability in downstream tasks. Spherical Geometry Diffusion addresses these challenges by producing high-quality explicit meshes through simple transformations, ensuring uniform mesh quality and superior geometric properties.

\section{Geometric Latent Representations}
\label{Supp:3}
In this section, we describe all the loss functions used in the VQVAE training process. The training of VQVAE involves two main components: \emph{Reconstruction} Loss and \emph{Geometric Regularization}, which will be discussed separately below.

\subsection{Reconstruction Loss}

The overall objective for reconstructing the Spherical Geometric Representation (SGR) is optimized within a Vector Quantized Variational Autoencoder (VQ-VAE) framework, which leverages both vector quantization and adversarial learning. The process begins by encoding an input SGR $\mathbf{G}$ with an encoder $E$ to obtain a latent representation $z = E(\mathbf{G})$. This vector is then quantized by mapping it to its nearest neighbor $e_k$ in a learned codebook $\mathcal{E}$, yielding $z' = \text{Quantize}(z)$. The resulting quantized latent $z'$ is then fed to a generator network $G$ to produce the reconstructed SGR $\hat{\mathbf{G}} = G(z')$.

The training objective is composed of four distinct loss terms. The primary component is the reconstruction loss $\mathcal{L}_{\text{rec}}$, which ensures the reconstructed SGR $\hat{\mathbf{G}}$ is faithful to the original $\mathbf{G}$ using a point-wise L2 distance:
$$
\mathcal{L}_{\text{rec}} = \frac{1}{H \times W} \sum_{i=1}^{H} \sum_{j=1}^{W} \left\| \mathbf{G}_{ij} - \hat{\mathbf{G}}_{ij} \right\|_2^2
$$
Additionally, to preserve high-level structural and semantic features, we employ a perceptual loss, $\mathcal{L}_{\text{per}}$. This loss compares feature maps extracted by a pre-trained network $\mathcal{F}$ (e.g., VGG \cite{simonyan2014very}) to measure the distance in a learned feature space:
$$
\mathcal{L}_{\text{per}} = \left\| \mathcal{F}(\mathbf{G}) - \mathcal{F}(\hat{\mathbf{G}}) \right\|_2^2
$$
To enhance realism, an adversarial loss $\mathcal{L}_{\text{adv}}$ drives the generator $G$ to produce SGRs that a discriminator $D$ finds indistinguishable from real data. The generator's objective is formulated as:
$$
\mathcal{L}_{\text{adv}} = - \mathbb{E}_{\mathbf{G} \sim p_{\text{data}}} [ \log D(\hat{\mathbf{G}}) ]
$$
Finally, we incorporate a mesh regularization loss, $\mathcal{L}_{\text{reg}}$, to encourage smoothness and prevent artifacts in the final derived mesh surface.

The total objective is a weighted sum of these components. To stabilize training, the adversarial component is introduced only after an initial phase of optimizing the other losses. The full loss is:
$$
\mathcal{L}_{\text{total}} = \lambda_{\text{rec}}\mathcal{L}_{\text{rec}} + \lambda_{\text{per}}\mathcal{L}_{\text{per}} + \lambda_{\text{reg}}\mathcal{L}_{\text{reg}} + \lambda_{\text{adv}}\mathcal{L}_{\text{adv}}
$$
where the $\lambda$ terms are weighting hyperparameters. Following the approach in Taming Transformers~\cite{esser2021taming}, we set the weight $\lambda_{\text{adv}} = 0.1$, $ \lambda_{\text{rec}}=\lambda_{\text{per}}=1$.

\begin{figure*}
    \centering
    \includegraphics[width=\textwidth]{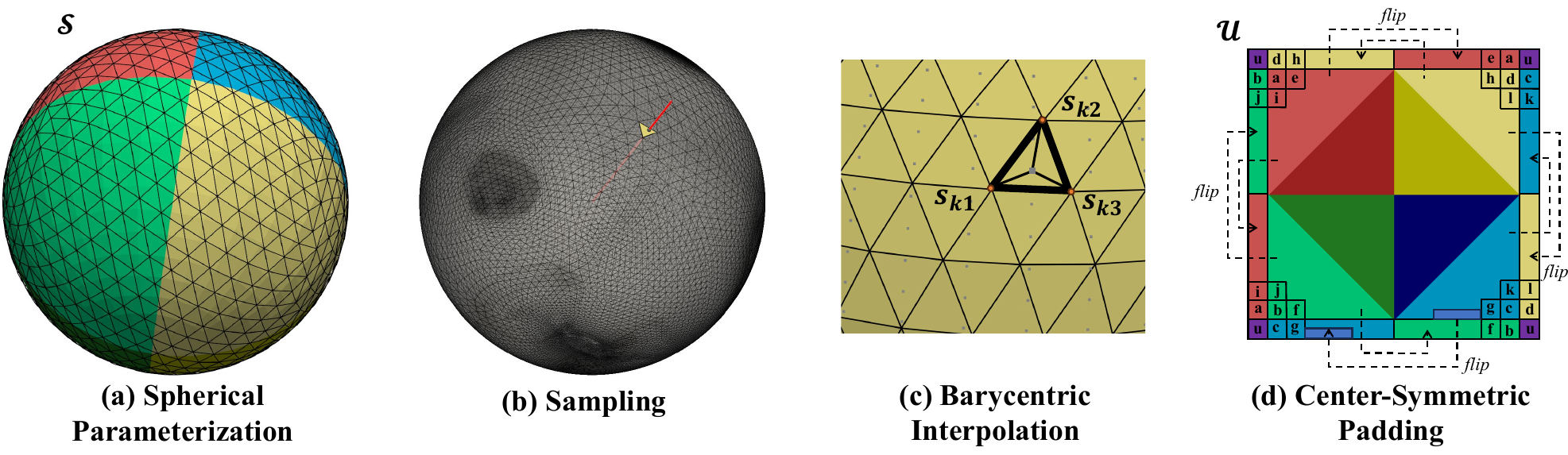} %
    \caption{\textbf{(a)} We obtain the spherical domain \( \mathcal{S} \) via Spherical Parameterization \( \mathcal{M} \rightarrow \mathcal{S}\), where \( \mathcal{U} \) in (d) denotes the unfold grid. Identical colors indicate corresponding regions. \textbf{(b)} and \textbf{(c)} illustrate the \emph{Uniform Sampling} process for a single sampling point, whose value is obtained via Barycentric Interpolation. \textbf{(d)} The proposed \emph{Center-symmetric Padding} mirrors the values symmetrically about the center of each edge. The purple color indicates averaging, with identical values marked by the same letters and colors.}
    \label{fig:spherical_s}
\end{figure*}
\subsection{Geometric Regularization.}
Given a reconstructed mesh \( M = (V, F) \), where \( V \) denotes the set of vertices and \( F \) represents the set of faces, we define the regularization term as a weighted sum of these losses.

\textbf{\textit{Normals consistency loss}}, \( \mathcal{L}_{nor} \), encourages the alignment of the normals of adjacent faces in the mesh. For two neighboring faces \( f_0 \) and \( f_1 \) with normals \( n_0 \) and \( n_1 \), respectively, the loss is computed as:
\begin{equation}
\mathcal{L}_{nor} = \sum_{\langle f_0, f_1 \rangle} \left( 1 - \frac{n_0 \cdot n_1}{\|n_0\| \|n_1\|} \right)
\end{equation}
where \( \cdot \) represents the dot product and \( \| \cdot \| \) denotes the Euclidean norm of the normals. This term minimizes angular discrepancies between normal vectors, ensuring smooth transitions between adjacent faces.

\textbf{\textit{Laplacian smoothing loss}}, \( \mathcal{L}_{lap} \), promotes smoothness throughout the mesh by penalizing large variations in vertex positions. It is defined as:
\begin{equation}
\mathcal{L}_{lap} = \sum_{i=1}^{N_v} \left\| L v_i \right\|^2
\end{equation}
where \( N_v \) is the number of vertices in the mesh, and \( L \) is the Laplacian matrix applied to vertex \( v_i \). This term encourages the vertices to stay close to their neighbors, leading to a smoother surface.

\textbf{\textit{Edge length regularization loss}}, \(\mathcal{L}_{edge}\), ensures that the edge lengths of the mesh are consistent. For each edge \( e \) in the mesh, the loss compares the actual edge length \( \| e \| \) to a target length \( e_0 \):
\begin{equation}
\mathcal{L}_{edge} = \frac{1}{N_m} \sum_{m=1}^{N_m} \frac{1}{E_m} \sum_{e \in m} \left( \| e \| - e_0 \right)^2
\end{equation}
where \( N_m \) is the number of meshes, \( E_m \) is the number of edges in each mesh, and \( \| e \| \) represents the length of edge \( e \). This term ensures that all edges of the mesh are approximately the same length, promoting uniformity and structural integrity.

The total \textit{Geometric Regularization} loss \( \mathcal{L}_{reg} \) is a weighted sum of three terms:
\begin{equation}
\mathcal{L}_{reg} = \alpha_{nor} \cdot \mathcal{L}_{nor} + \alpha_{lap} \cdot \mathcal{L}_{lap} + \alpha_{edg} \cdot \mathcal{L}_{edge}
\end{equation}
where \( \alpha_{nor}, \alpha_{lap}, \alpha_{edg} \) are the respective weights for each loss term, allowing for controlled emphasis on different aspects of \textit{Geometric Regularization}. In Geometric Regularization, we set \( \alpha_{\text{nor}} = 0.1 \), \( \alpha_{\text{lap}} = 0.5 \), and \( \alpha_{\text{edg}} = 0.1 \), respectively.

\subsection{Center-symmetric Padding.}
When the SGR is unwrapped in a 2D image, it inherently retains spherical continuity. We observed that directly applying traditional zero padding leads to cracks in 3D shapes. To address this issue, we propose a new padding, \emph{Center-symmetric Padding}, to better adapt to the spherical continuity of the SGR.
Specifically, as illustrated in Fig.~\ref{fig:spherical_s},each edge of \( \mathcal{U} \) is centrally symmetric in \( \mathcal{S} \), and the four corners of \( \mathcal{U} \) are averaged in \( \mathcal{S} \). The pseudocode for \textit{Center-symmetric Padding} is shown in Fig.~\ref{alg:Center-symmetric Padding}. 
We use centrally-symmetric pixels as outer padding to convey 3D connectivity, thereby mitigating discontinuities.
\begin{figure}[h]
    \centering
    \includegraphics[scale=0.29, keepaspectratio]{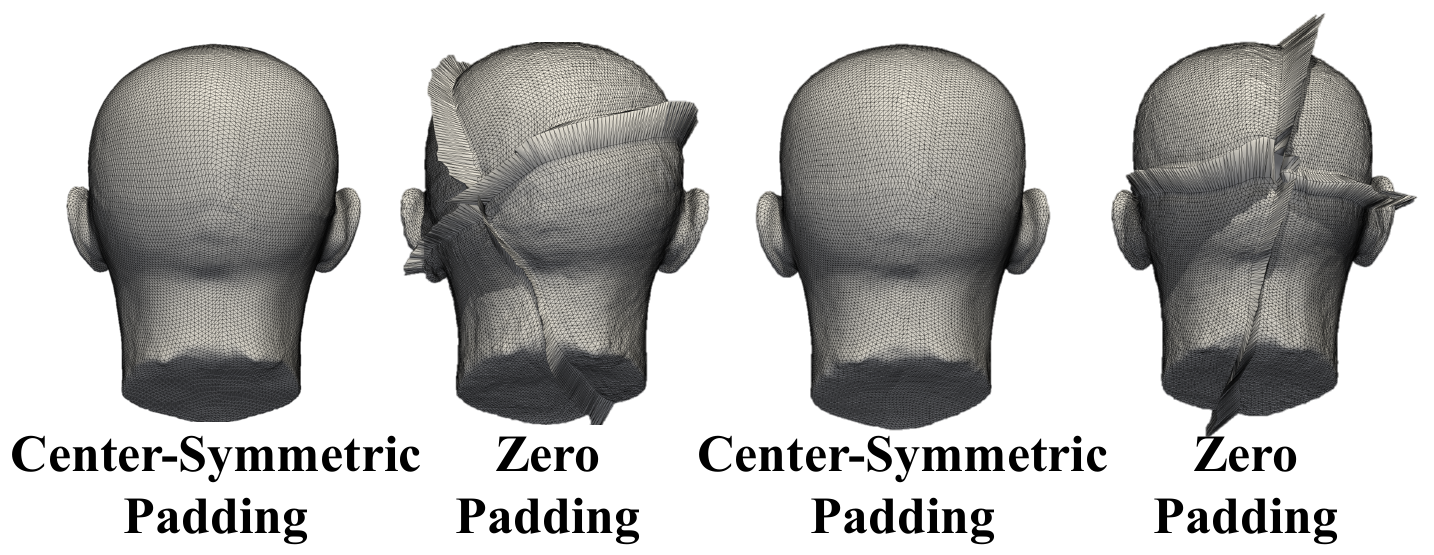} 
    \caption{Impact of \emph{Center-symmetric Padding}.} %
    \label{fig:padding_cut_s}
\end{figure}
\begin{algorithm}[t]
\caption{Center-symmetric Padding}
\label{alg:Center-symmetric Padding}
\begin{algorithmic}[1]
\Require 
  Input $\mathbf{G} \in \mathbb{R}^{H \times W}$
\Ensure 
  Padded output $\mathbf{Q} \in \mathbb{R}^{(H+2) \times (W+2)}$ 
\State $\mathbf{Q}_{1:H, 1:W} \gets \mathbf{G}_{0:H-1, 0:W-1}$ \Comment{Copy center block}
\For{$i \gets 1$ \textbf{to} $H$}
  \State $\mathbf{Q}_{i,0} \gets \mathbf{G}_{H-i,0}$ \Comment{Left flip}
  \State $\mathbf{Q}_{i,W+1} \gets \mathbf{G}_{H-i,W-1}$ \Comment{Right flip}
\EndFor
\For{$j \gets 1$ \textbf{to} $W$}
  \State $\mathbf{Q}_{0,j} \gets \mathbf{G}_{0,W-j}$ \Comment{Top flip}
  \State $\mathbf{Q}_{H+1,j} \gets \mathbf{G}_{H-1,W-j}$ \Comment{Bottom flip}
\EndFor
\State $\nu \gets \frac{1}{4}(\mathbf{G}_{0,0} + \mathbf{G}_{0,W-1} + \mathbf{G}_{H-1,0} + \mathbf{G}_{H-1,W-1})$
\State $\mathbf{Q}_{0,0}, \mathbf{Q}_{0,W+1}, \mathbf{Q}_{H+1,0}, \mathbf{Q}_{H+1,W+1} \gets \nu$
\State \Return $\mathbf{Q}$
\end{algorithmic}
\end{algorithm}

\section{More Implementation Details}
\label{Supp:5}
\paragraph{Mesh Preprocessing.}
Spherical parameterization imposes strict topological constraints, requiring input meshes to be, for instance, genus-zero manifolds. To meet these prerequisites, we preprocess our face meshes using the meshfix library~\cite{attene2010lightweight}. meshfix converts each input into a single, watertight triangle mesh that bounds a polyhedron by removing all singularities, self-intersections, and degenerate elements, while leaving defect-free regions of the surface unmodified. Crucially, during our subsequent SGR sampling stage, we flag all vertices introduced by this preprocessing step. This allows us to optionally discard these auxiliary vertices later, ensuring that modifications can be confined to the original mesh structure when necessary.

\paragraph{Selection of Spherical Mapping.}
Spherical parameterization is a well-established field offering methods categorized by their optimization objectives, including conformal, area-preserving, and low-distortion approaches. Our work targets high-quality mesh reconstruction, which requires generating elements that are simultaneously well-shaped and uniformly sized. This dual requirement for element quality and size uniformity mandates a careful selection among these parameterization strategies.

Neither of the classical approaches is suitable for this task. Although conformal maps perfectly preserve angles, thus ensuring well-shaped elements, they introduce severe area distortion. Consequently, a uniform grid in the parameter domain maps to a mesh of highly nonuniform density on the 3D surface, violating our goal of size uniformity. In contrast, area-preserving maps maintain uniform element sizing, but at the cost of significant angular distortion. This trade-off can transform equilateral triangles from the parameter domain into elongated, degenerate triangles on the 3D model, which undermines the core requirement for element quality. Clearly, methods that are inherently biased towards a single metric are ill-suited for our task, as any severe distortion in either angle or area is unacceptable.

Therefore, we adopt a low distortion spherical parameterization method~\cite{praun2003spherical}. The principle of such methods is to find an optimal trade-off between the preservation of angle and area. Instead of perfecting one metric, they minimized a unified distortion energy that penalizes both angular and areal deviation, effectively making the map ``as-isometric-as-possible." By using such a parameterization, when we map a uniform, high-quality triangulation from the sphere back to the 3D model, the low-distortion property ensures that the resulting elements are both well-proportioned and uniformly sized. This balanced approach provides a robust foundation for high-quality reconstruction, yielding results superior to those achievable with purely conformal or area-preserving methods.

\paragraph{Geometric Regularization Hyperparameters.}
\label{Geometric Regularization Hyperparameters.}
To validate the effectiveness of our proposed weights (\( \alpha_{\text{nor}} = 0.1 \), \( \alpha_{\text{lap}} = 0.5 \), \( \alpha_{\text{edg}} = 0.1 \)), we conducted a series of ablation studies. For a fair comparison, we tested different weight configurations and retrained the VQVAE model for 100 epochs for each setup. The quantitative results are presented in Table~\ref{tab:ablation_weights}. As shown in the table, our combination achieves the best results in the quantitative metrics, confirming the rationale behind our selection.

\begin{table}[htbp] %
  \centering
  \small
  \setlength{\tabcolsep}{10pt} %
  \begin{tabular}{ccccc}
    \toprule
    {\( \alpha_{\text{nor}}\)} & {\( \alpha_{\text{lap}}\)} & {\( \alpha_{\text{edg}}\)} & {CD (mm)$\downarrow$} & {F-score@1mm$\uparrow$} \\
    \midrule
    0.1 & 0.1 & 0.1 & 1.25  & 49.86 \\
    0.1 & 0.1 & 0.5 & 1.25  & 52.36 \\
    0.1 & 0.1 & 1   & \underline{1.16}  & \underline{55.72} \\
    0.5 & 0.1 & 0.1 & 1.28  & 47.48 \\
    0.5 & 0.1 & 0.5 & 1.17  & 54.24 \\
    0.5 & 0.1 & 1   & 1.17  & 54.44 \\
    0.5 & 0.5 & 0.5 & 1.26  & 49.21 \\
    0.5 & 1   & 0.5 & 1.29  & 48.50  \\
    1   & 1   & 1   & 1.34  & 47.69 \\
    0   & 0.5 & 0.1 & 1.23 & 53.50 \\
    0.1 & 0.5 & 0   & 1.27  & 47.12 \\
    0.1 & 1   & 0.1 & 1.45  & 47.78 \\
    0.1 & 0.5 & 0.1 & \textbf{1.04}   & \textbf{59.95}   \\
    \bottomrule
  \end{tabular}
  \caption{Ablation study on the loss weights.}
  \label{tab:ablation_weights}
\end{table}

\paragraph{Techniques}
We use Diffusers \cite{von-platen-etal-2022-diffusers} to implement the code, as it is a widely used diffusion framework capable of efficiently achieving high-quality diffusion generation. For mesh rendering, we employ PyVista as a visualization tool, which provides powerful functionalities for visualization.

\section{Experiments}
\label{Supp:4}

\subsection{3D Geometric Latent Representations}
\begin{figure*}
    \centering
    \includegraphics[width=\textwidth, height=0.9\textheight, keepaspectratio]{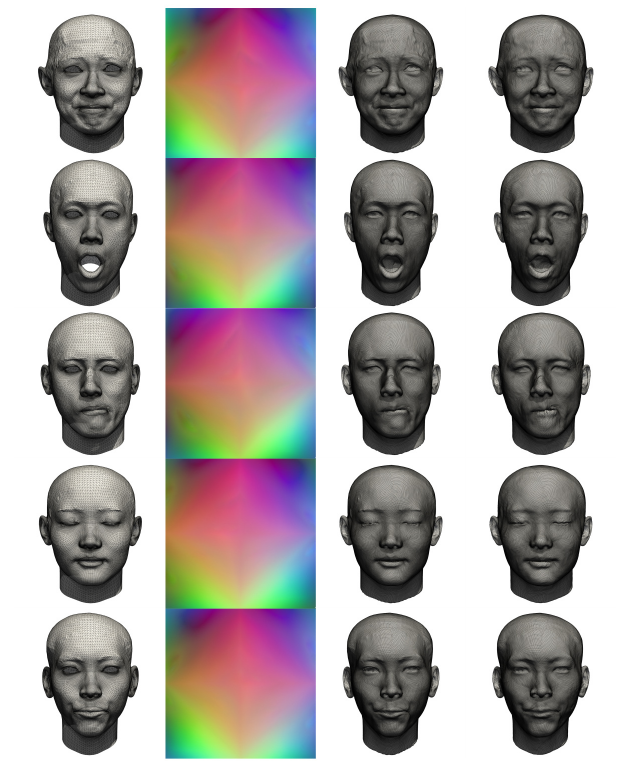} %
    \caption{Qualitative results of \textbf{\textit{Geometric Latent Representations}}. Each row, from left to right, shows the \textbf{ground truth}, \textbf{SGR}, the mesh \textbf{reconstructed from the SGR}, and the mesh \textbf{reconstructed by VQVAE}. Each row corresponds to different individuals exhibiting varying expressions.}
    \label{fig:sampels1}
\end{figure*}

\begin{figure*}
    \centering
    \includegraphics[width=\textwidth, height=0.9\textheight, keepaspectratio]{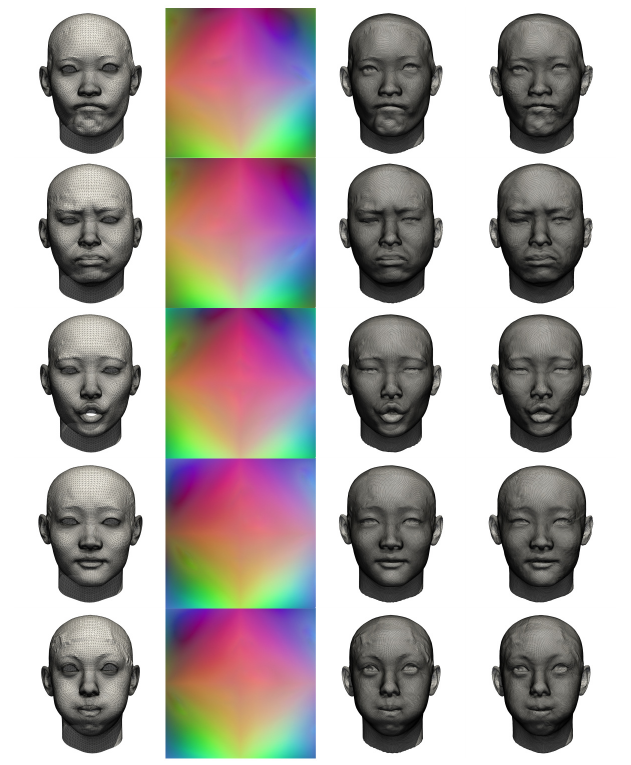} %
    \caption{Qualitative results of \textbf{\textit{Geometric Latent Representations}}. Each row, from left to right, shows the \textbf{ground truth}, \textbf{SGR}, the mesh \textbf{reconstructed from the SGR}, and the mesh \textbf{reconstructed by VQVAE}. Each row corresponds to different individuals exhibiting varying expressions.}
    \label{fig:sampels2}
\end{figure*}

To demonstrate the effect of VQVAE, we conducted a qualitative evaluation on the test set. During the evaluation process, the facial scans were first converted into SGR and then reconstructed using the VQVAE model. The qualitative results are presented in Fig.~\ref{fig:sampels1} and Fig.~\ref{fig:sampels2}. As demonstrated, representation learning is highly effective, accurately reconstructing explicit meshes for unseen test data.

\subsection{Conditional Face Generation}
\label{Conditional Face Generation}
\begin{figure*}
    \centering
    \includegraphics[width=0.9\textwidth]{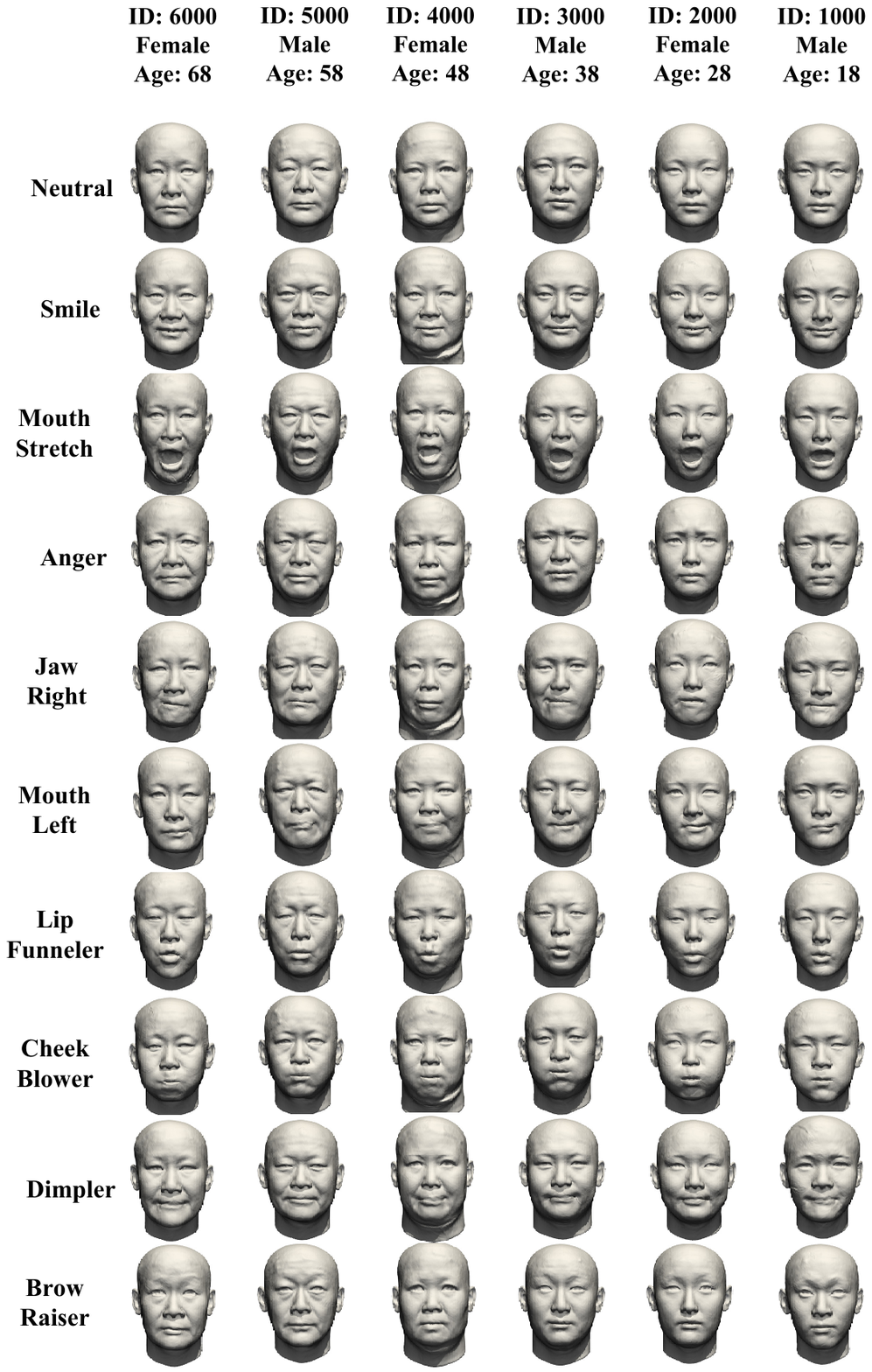}
    \caption{More qualitative results of \textbf{\textit{Conditional Face Generation}}.}
    \label{fig:4_more}
\end{figure*}

We present additional \emph{Conditional Face Generation} results. Fig.~\ref{fig:4_more} showcases additional results across a diverse range of identities, expressions, and ages. As demonstrated, SGR effectively captures identity information, generating high-quality meshes that faithfully represent the facial features of the input text, while the correspondence between identities is maintained across different expressions.

\noindent \textbf{Quantitative Evaluation.} 
In this section, we quantitatively evaluate the effect of different \textit{inference steps} and \textit{Guidance Scales} on the quality of generation.
Consistent with previous research~\cite{abrevaya2019decoupled,taherkhani2023controllable}, the evaluation focuses on two key aspects: the \emph{Diversity} and \emph{Specificity} of the generated 3D shapes.

\emph{Diversity:} A crucial metric for evaluating generative models is the diversity of the generated samples. This is measured by calculating the average vertex distance between \( n_1 \) generated samples. A higher diversity value indicates that the model can produce a broader range of distinct samples. In this experiment, we set \( n_1 = 10000 \).

\emph{Specificity:}  
Specificity quantifies how closely the generated shapes align with the distribution of the original training data, with the aim of ensuring that the generated data closely mirror the original data distribution. To assess specificity, we randomly generate \( n_2 \) samples and compute the average vertex distance between each generated sample and all training samples in the dataset. The minimum distance obtained is considered the distance between the generated sample and the original data distribution. In this experiment, \( n_2 = 1000 \). The comparison results using these metrics are presented in Table~\ref{tab:1}. 

It is important to note that as the guidance scale increases, the generated faces exhibit greater specificity. This leads to an increase in the average distance between 1000 pairs of different conditions, resulting in higher Diversity values.

\subsection{Conditional Texture Generation.}
To facilitate texture generation, we first convert the general text descriptions into text prompts that specifically describe the texture. This process is achieved using GPT and a predefined template, an approach inspired by InstructPix2Pix~\cite{brooks2023instructpix2pix}. The template we use is presented in Listing. \ref{lst:prompt_template}.
\begin{lstlisting}[
    caption={Template for Instruct-pix2pix Prompt},
    label={lst:prompt_template}
    ]
You are an expert in 3D graphics and texture generation. Your task is to convert detailed geometric and appearance descriptions into concise texture generation instructions suitable for instruct_pix2pix.

## Input Format:
A detailed description containing both geometric features (shape, structure) and appearance features (color, texture, material properties).

## Output Format:
Generate a concise instruction in the style of instruct_pix2pix that focuses on texture and surface appearance, following this pattern:
"Convert this 3D geometry to [facial texture description] with [specific surface details]"

## Guidelines:
1. Focus on SURFACE APPEARANCE
2. Extract color, texture, and material information
3. Keep the instruction concise but descriptive
4. Include relevant surface details (smooth, rough, aged, glossy, etc.)
5. IMPORTANT: Return ONLY the instruction text, no explanations, no additional text, no quotes

## Examples:

Input: "She has no beard. She has wide mouth and bow-shaped, thin lips. Her nose is big, wide and upturned with a high nose bridge. Her face is square and thin. This old woman is Westerner. Her eyes are medium sized, triangle and brown with double eyelid. She has medium width eye distance. Her eyebrow is round, sparse and black."

Output: "Convert this 3D facial geometry to realistic aged Caucasian facial texture with natural wrinkles, brown eyes, and sparse black eyebrows."

## Task:
Convert the following description into an instruct_pix2pix compatible instruction:

Input: "{description}"

Output:
\end{lstlisting}

\begin{table}[t]  
  \centering  
  \small
  \setlength{\tabcolsep}{4pt}
\begin{tabular}{lcccc}
    \toprule
    \textbf{Method} & \textbf{DIV} & \textbf{DIV-ID} & \textbf{DIV-EXP} & \textbf{SP (mm)} \\
    \midrule
    Training data           & 1.00 & 1.00 & 1.00 & -    \\
    \hline %
    3DMM & 0.72 & 0.59 & 0.57 & 2.30 \\
    MAE & 0.79 & 0.28 & 0.75 & 2.00 \\
    CoMA & 0.69 & 0.52 & 0.58 & 2.47 \\
    FacialGAN & 0.96 & 0.58 & 0.84 & 2.01 \\
    D3FSM & 0.77 & 0.81 & 0.37 & 0.84 \\
    \midrule
    \textbf{Inference Steps = 20} &      &      &      &      \\
    GS = 1                  & \textbf{0.83} & 0.88 & 1.60 & \textbf{0.82} \\
    GS = 4                  & 1.21 & 1.28 & 2.31 & 3.10 \\
    GS = 7                  & 1.91 & 2.03 & 3.62 & 5.32 \\
    \midrule
    \textbf{Inference Steps = 30} &      &      &      &      \\
    GS = 1                  & \textbf{0.95} & 0.87 & 1.58 & \textbf{0.82} \\
    GS = 4                  & 1.12 & 1.00 & 1.80 & 1.01 \\
    GS = 7                  & 1.12 & 1.19 & 2.15 & 2.10 \\
    \midrule
    \textbf{Inference Steps = 40} &      &      &      &      \\
    GS = 1                  & \textbf{0.93} & 0.88 & 1.58 & \textbf{0.83} \\
    GS = 4                  & 0.93 & 0.99 & 1.80 & 0.96 \\
    GS = 7                  & 1.03 & 1.10 & 2.15 & 1.59 \\
    \bottomrule
\end{tabular}
\caption{Quantitative metrics w.r.t normalized Diversity (\textbf{DIV}, \textbf{DIV-ID}, \textbf{DIV-EXP}) and absolute \textbf{Specificity} (\textbf{SP}). \textbf{SP} is measured in millimeters (lower is better). \textbf{DIV}, \textbf{DIV-ID}, and \textbf{DIV-EXP} represent diversity metrics (higher is better). \textbf{Inference Steps} are evaluated with different \textbf{Guidance Scales} (\textbf{GS}).}
\label{tab:1}
\end{table}

\begin{figure*}
    \centering
    \includegraphics[width=0.9\textwidth]{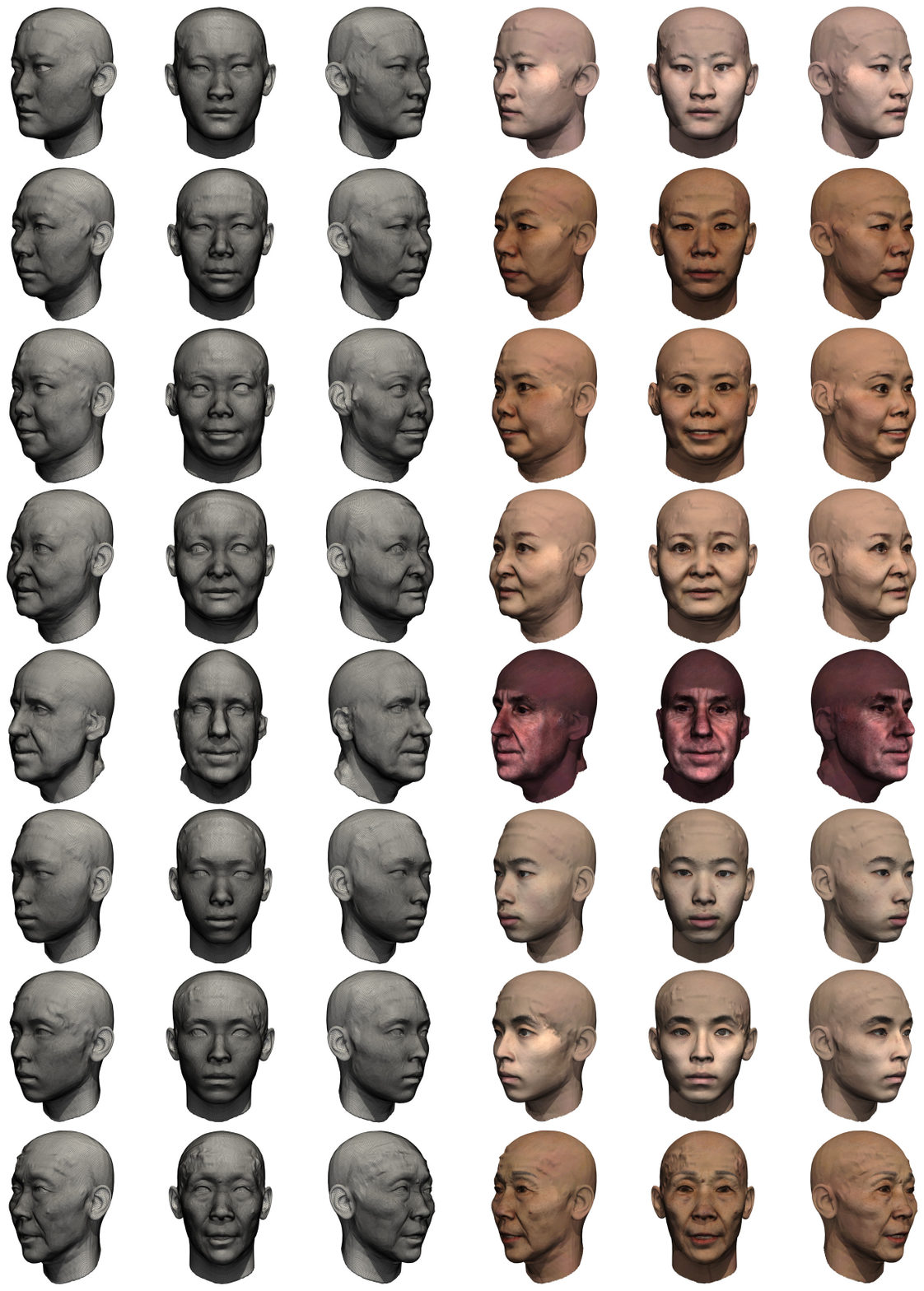}
    \caption{More qualitative results of \textbf{\textit{Conditional Texture Generation.}}.}
    \label{fig:texture_more1}
\end{figure*}

\begin{figure*}
    \centering
    \includegraphics[width=0.9\textwidth]{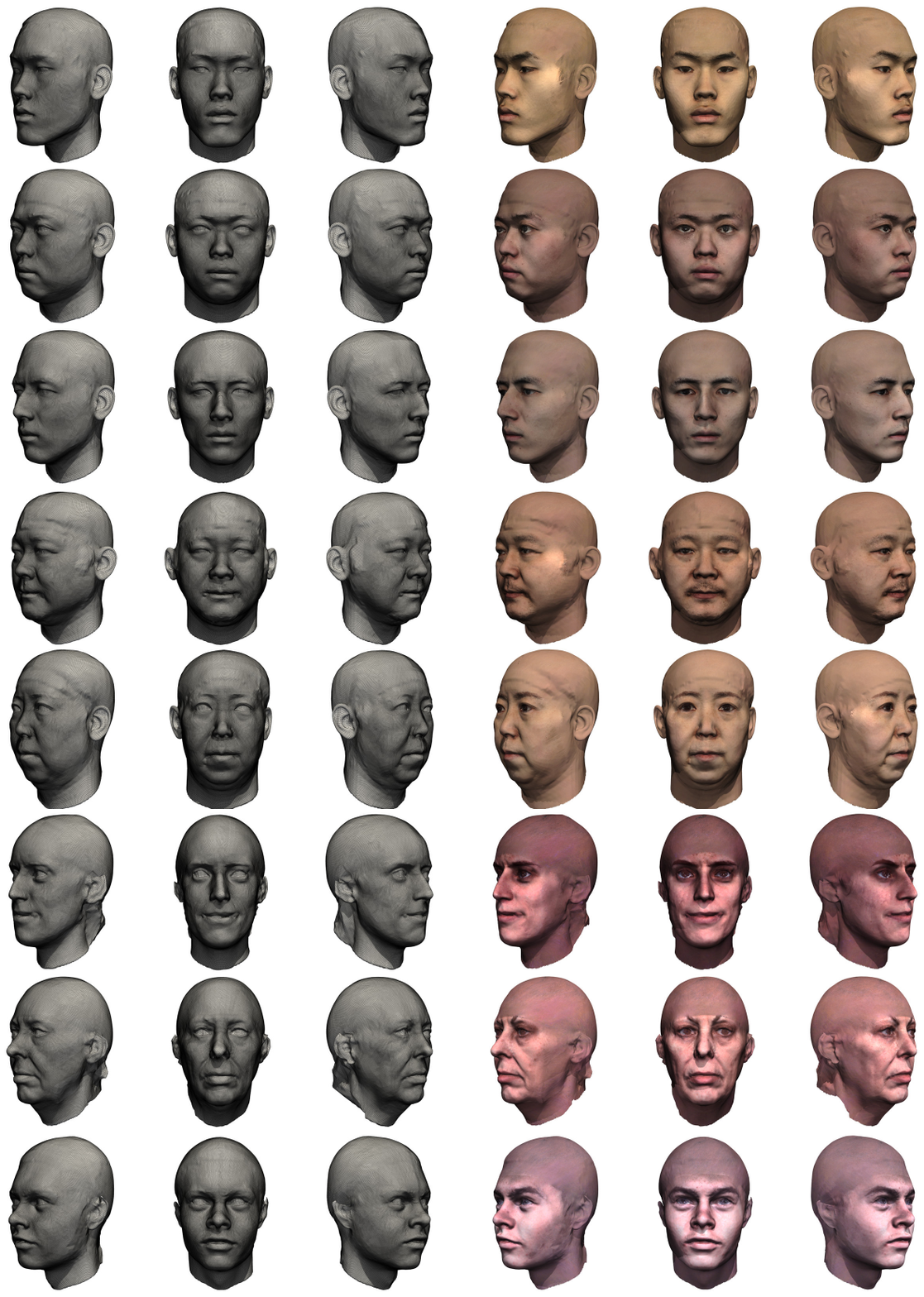}
    \caption{More qualitative results of \textbf{\textit{Conditional Texture Generation.}}.}
    \label{fig:texture_more2}
\end{figure*}

\noindent \textbf{Qualitative results.} 
We present additional results for Conditional Texture Generation. Fig.~\ref{fig:texture_more1} and Fig.~\ref{fig:texture_more2} showcases results across a diverse range of viewpoints and text prompts, presenting both the underlying geometric renderings and the final textured outputs. These results visually demonstrate the effectiveness of our method in generating diverse textures.

\subsection{Conditional Face Reconstruction}
We present additional \emph{Conditional Face Reconstruction} results based on SGR. The results are shown in Fig.~\ref{fig:mesh quality2} and Fig.~\ref{fig:mesh quality1}. 
As demonstrated, SGR outperforms the state-of-the-art implicit function method ImFace++ \cite{zheng2024imface++} in terms of geometric accuracy and quality. SGR produces explicit meshes with higher geometric accuracy and quality, particularly in terms of mesh quality, where no distorted triangles are present.

\noindent \textbf{Quantitative Evaluation.} 
We quantitatively evaluate the impact of different SGR \textit{resolutions} and numbers of \textit{inversion steps} on reconstruction time. The quantitative results are presented in Table~\ref{table:combined}.
\begin{table}[ht]  
  \centering  
    \begin{tabular}{l|cccc}  
      \toprule  
      Steps & $R=32$ & $R=64$ & $R=128$ & $R=256$ \\
      \midrule  
      100 & $\sim$2.83\,s & $\sim$2.87\,s & $\sim$2.95\,s & $\sim$3.80\,s \\
      300 & $\sim$8.28\,s & $\sim$8.36\,s & $\sim$8.47\,s & $\sim$11.15\,s \\
      600 & $\sim$16.55\,s & $\sim$16.72\,s & $\sim$16.83\,s & $\sim$22.35\,s \\
      \bottomrule  
    \end{tabular}  
      \caption{The inversion computation time for various resolutions $R$ and inversion steps.}  
      \label{table:combined}  
\end{table}

\begin{figure*}[t]
    \centering
    \includegraphics[scale=0.8, keepaspectratio]{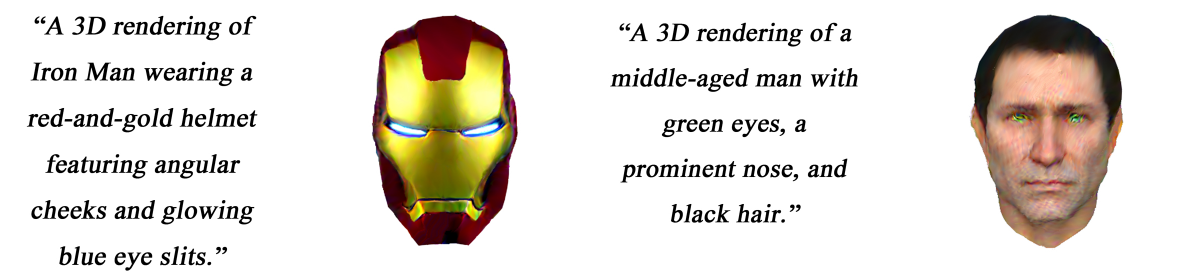} 
    \caption{SDS-based synthesis results. Detailed descriptions and SGR's 3D renderings.}
    \label{fig:rebuttal_3}
\end{figure*}

\subsection{Geometric quality}
\begin{figure*}
    \centering
    \includegraphics[width=\textwidth, height=0.9\textheight, keepaspectratio]{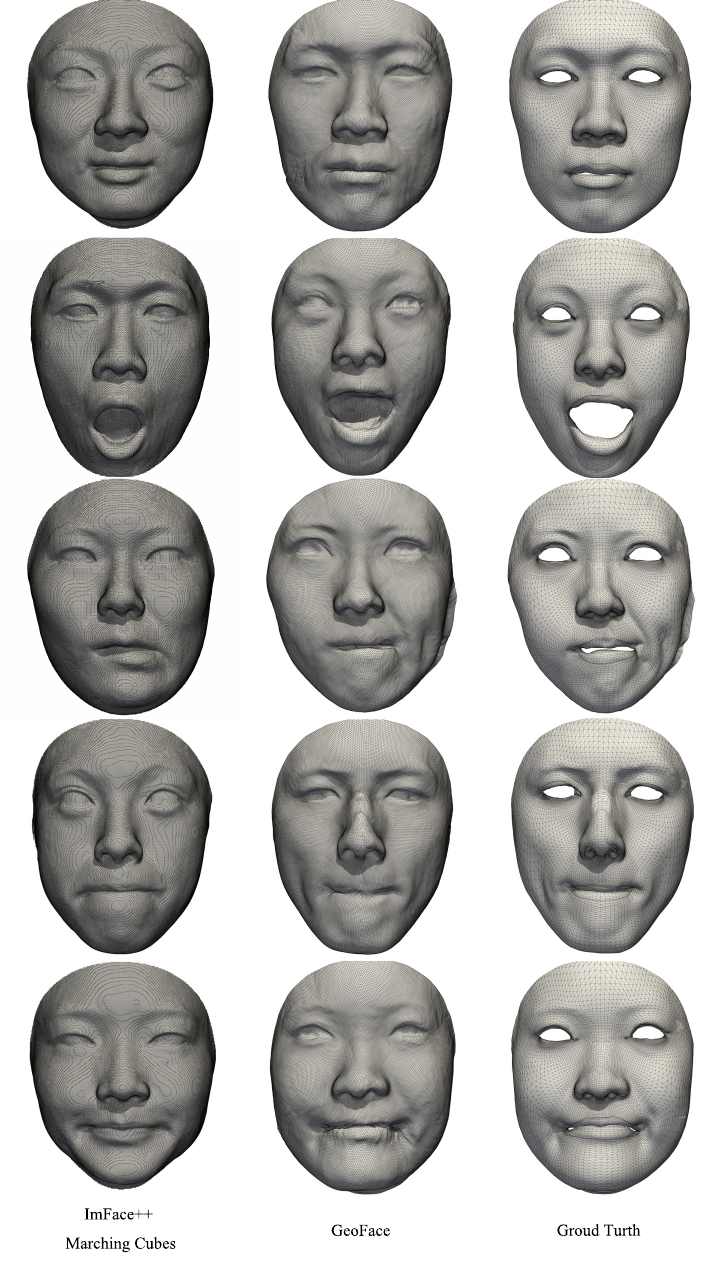} %
    \caption{Qualitative comparison of \textbf{mesh quality} and \textbf{\textit{Conditional Face Reconstruction}}. The comparison is between meshes generated by the Marching Cubes algorithm and those produced by Spherical Geometric Diffusion, where the Marching Cubes meshes are derived from the results of ImFace++.}
    \label{fig:mesh quality2}
\end{figure*}

\begin{figure*}
    \centering
    \includegraphics[width=\textwidth, height=0.9\textheight, keepaspectratio]{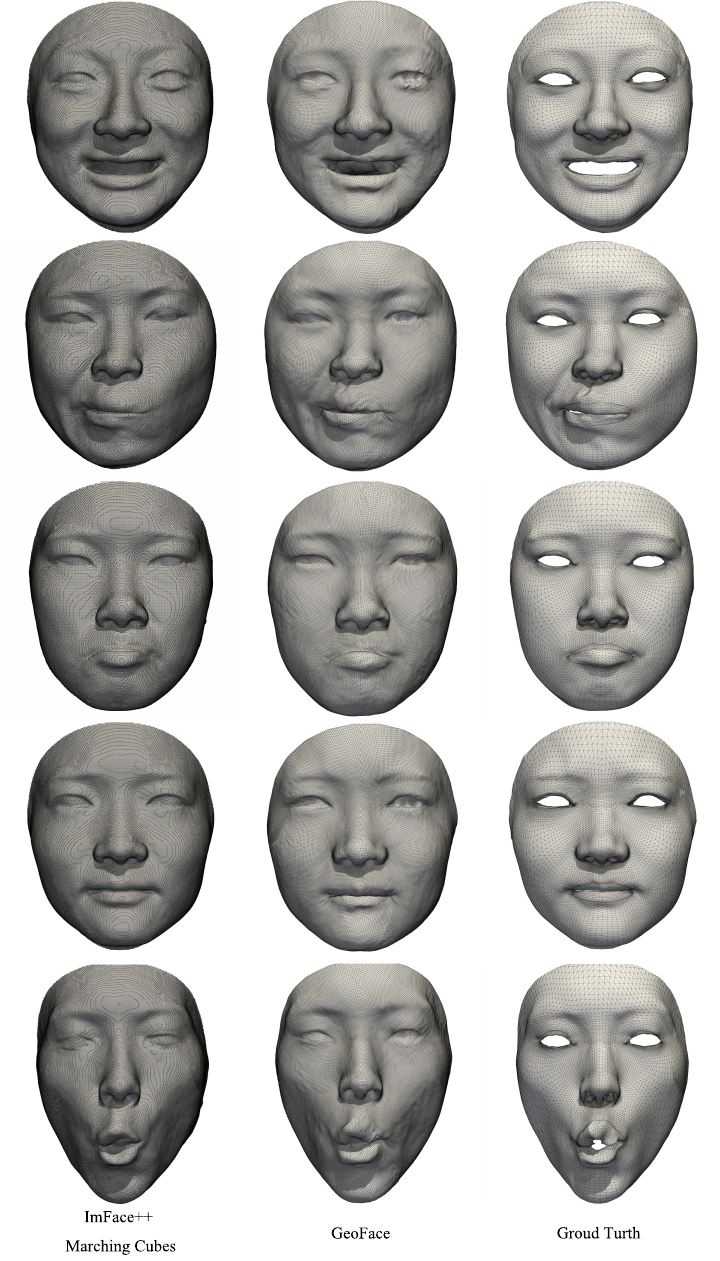} %
    \caption{Qualitative comparison of \textbf{mesh quality} and \textbf{\textit{Conditional Face Reconstruction}}. The comparison is between meshes generated by the Marching Cubes algorithm and those produced by Spherical Geometric Diffusion, where the Marching Cubes meshes are derived from the results of ImFace++.}
    \label{fig:mesh quality1}
\end{figure*}

We present additional SGR results and compare the mesh quality quantitatively and qualitatively with the Marching Cubes algorithm~\cite{lorensen1998marching}, which is a widely used method for reconstructing meshes from implicit functions. The qualitative results are shown in Fig.~\ref{fig:mesh quality2} and Fig.~\ref{fig:mesh quality1} , where it is evident that SGR produces more uniform explicit meshes, while the meshes generated by Marching Cubes exhibit noticeable optimization artifacts. Additionally, SGR demonstrates higher geometric accuracy, capturing finer details, and producing results closer to the ground truth. %

Aspect ratio is a commonly used metric to evaluate the shape quality of mesh elements. An aspect ratio closer to 1 indicates that the element is more regular in shape, such as an equilateral triangle. This metric is significant for identifying distorted or poorly shaped elements, which can adversely affect the stability and accuracy of numerical simulations in applications such as 3D geometry processing. The quality criterion for the aspect ratio of a triangular mesh element is defined to assess the conformity of the element to an equilateral triangle, where all edges are of equal length. The aspect ratio \( q \) is calculated as follows:

\[
q = \frac{L_{\text{max}} (L_0 + L_1 + L_2)}{4 \sqrt{3} A},
\]
where \( L_0 \), \( L_1 \), and \( L_2 \) represent the lengths of the three edges of the triangle, and \( L_{\text{max}} \) denotes the maximum edge length:

\[
L_{\text{max}} = \max(L_0, L_1, L_2).
\]
Area \( A \) of the triangle is calculated using the cross product of two edge vectors:
\[
A = \frac{1}{2} \| \mathbf{L}_0 \times \mathbf{L}_1 \|,
\]
where \( \mathbf{L}_0 \) and \( \mathbf{L}_1 \) are vectors representing two sides of the triangle. This metric ensures that the aspect ratio reflects the deviation of the triangle's geometry from an ideal equilateral triangle. Lower values of \( q \) indicate better geometric quality. 

The quantitative results are shown in the Table. 2 of the main paper. As demonstrated, SGR significantly outperforms other methods in terms of mesh quality.

\subsection{SDS-based  synthesis}
Privacy concerns restrict large-scale datasets that pair detailed descriptions with 3D facial meshes. Current detailed text-driven face synthesis methods rely on Score Distillation Sampling (SDS)~\cite{poole2022dreamfusion} optimization, which can also be applied to SGR. As illustrated in Fig.~\ref{fig:rebuttal_3}, we optimize the geometry and texture SGR using SDS to capture detailed text-driven features such as eyes and nose.

\subsection{Ablation Study}
\noindent \textbf{Geometric Latent Representations.}
To quantitatively compare the impact of \textit{Geometric Regularization}, we conducted a preliminary experiment using VQVAE without Geometric Regularization as the baseline. The results are shown in the Table. 4 of the paper. We refer to the VQVAE with mesh quality regularization as \textit{VQVAE-mesh} and the one without it as \textit{VQVAE-base}. For a fair comparision, both models were trained for 30,000 steps.  As indicated by the results, \textit{Geometric Regularization} penalizes deformed mesh outputs, enabling VQVAE to converge more quickly during the early stages of learning.

We also investigated the impact of different weight parameters on Geometric Regularization. Specific results and analysis are presented in Table~\ref{tab:ablation_weights}.

\smallskip
\noindent \textbf{Conditional Face Reconstruction.}
We observed that in DDIM-based inversion for reconstructing new face scans, the reconstruction quality is dependent on the number of inversion steps. Typically, the greater the number of inversion steps, the higher the reconstruction quality, though this comes with a greater time cost. As shown in Table. 4 of the paper, we performed quantitative experiments to assess the impact of inversion steps on reconstruction quality. We found that as the number of inversion steps increases, both reconstruction metrics improve. However, after a certain number of steps, further increases in the inversion steps yield negligible improvements in reconstruction quality. 

The number of inversion steps and the resolution also affect the computational time. We conducted an ablation study to investigate this, and the corresponding results are presented in Table~\ref{table:combined}.

\smallskip
\noindent \textbf{Conditional Face Generation.}
The number of inference steps and the guidance scale are known to affect the generation quality of diffusion models. We conducted an ablation study to analyze these effects, with the results and analysis presented in Table~\ref{tab:1}.

\clearpage

